\pdfoutput=1
\documentclass[11pt]{article}

\usepackage[]{acl}

\usepackage{times}
\usepackage{latexsym}

\usepackage[T1]{fontenc}
\usepackage[utf8]{inputenc}

\usepackage{microtype}

\usepackage{graphicx} % for all sorts of formatting
\usepackage{multirow} % for table formatting
\usepackage{booktabs} % for table formatting
\usepackage{multirow} % for table formatting
\usepackage{subcaption} % table subfigures
\usepackage{tikz}
\newlength\barheight 
\newlength\barwidth 
\newcommand{\DrawPercentageBar}[1]{%
\centering
  #1\%
  \begin{tikzpicture}
    \fill[color=black!10] (-\barwidth, -\barheight) rectangle (0.0, 0.0);
    \fill[color=black]   (-#1/100*\barwidth , -\barheight ) rectangle (0.0 , 0.0);
  \end{tikzpicture}%
}

\DeclareUnicodeCharacter{2014}{\dash}

\newcommand\blfootnote[1]{%
  \begingroup
  \renewcommand\thefootnote{}\footnote{#1}%
  \addtocounter{footnote}{-1}%
  \endgroup
}

\usepackage{makecell}
\usepackage{listings}
\usepackage{times,latexsym}
\usepackage{url}
\usepackage{color,soul} % for \hl{}
\usepackage[T1]{fontenc}
\usepackage{amsmath}

\usepackage{array,booktabs,ragged2e}
\usepackage{multirow}
\newcolumntype{R}[1]{>{\RaggedLeft\arraybackslash}p{#1}}
\newcolumntype{L}[1]{>{\RaggedRight\arraybackslash}m{#1}}% Added by Caleb
\usepackage{pifont} % Added by Caleb

\newcommand{\Data}[1]{\textsc{Positive Psychology Frames}{#1}}
\newcommand{\numDatapoints}{8,349}
\newcommand{\numStructuredAnnotations}{12,755}
\newcommand{\numReframingStrategies}{six}

\newcommand{\numWorkers}{204}
\newcommand{\numHumanEvals}{50}

\newcommand{\Meaning}[1]{\textcolor{violet}{#1}}
\newcommand{\Positivity}[1]{\textcolor{teal}{#1}} 
\newcommand{\Fluency}[1]{\textcolor{brown}{#1}} 

\newcommand{\best}[1]{\colorbox{black!10}{#1}}

\usepackage{amsmath, balance}
\usepackage{caption}
\usepackage{subcaption}
\usepackage{booktabs}
\usepackage{soul}

\newcommand\coauth{$^\star$}
\newcommand{\gt}{$^\dagger$}
\newcommand{\nus}{$^\diamond$}

\title{Inducing Positive Perspectives with Text Reframing}

\author{Caleb Ziems \coauth \gt \hspace{1.5em}
        Minzhi Li \coauth \nus \hspace{1.5em}
        Anthony Zhang \gt \hspace{1.5em}
        Diyi Yang \gt \hspace{1.5em} \\
        \gt Georgia Institute of Technology\\
        \texttt{\{\href{mailto://cziems3@gatech.edu}{cziems}, \href{mailto://azhang305@gatech.edu}{azhang305}, \href{mailto://dyang888@gatech.edu}{dyang888}\}@gatech.edu} \\
        \nus National University of Singapore \\
        \texttt{\href{mailto://li.minzhi@u.nus.edu}{li.minzhi@u.nus.edu}}
}

\date{}

\begin{document}
\maketitle
\begin{abstract}
Sentiment transfer is one popular example of a text style transfer task, where the goal is to reverse the sentiment polarity of a text. With a sentiment reversal comes also a reversal in meaning. We introduce a different but related task called \textit{positive reframing} in which we neutralize a negative point of view and generate a more positive perspective for the author without contradicting the original meaning. Our insistence on meaning preservation makes positive reframing a challenging and semantically rich task. To facilitate rapid progress, we introduce a large-scale benchmark, \Data, with \numDatapoints{} sentence pairs and \numStructuredAnnotations{} structured annotations to explain positive reframing in terms of six theoretically-motivated reframing strategies. Then we evaluate a set of state-of-the-art text style transfer models, and conclude by discussing key challenges and directions for future work. To download the data, see \url{https://github.com/GT-SALT/positive-frames}\blfootnote{\coauth Equal contribution.}
\end{abstract}
\section{Introduction}
\begin{quote}
    {
    Gratitude is not only the greatest of virtues, but the parent of all the others. \\\phantom{abcdefg}--- \textbf{Marcus Tullius Cicero}
    }
\end{quote}

Text style transfer (TST) has received much attention from the language technologies community \cite{hovy1987generating,jin2020deep}, where the goal is to change some attribute, like the sentiment of the text, without changing any attribute-independent content \cite{mir2019evaluating,fu2018style,logeswaran2018content}. 
Some TST applications such as de-biasing \cite{pryzant2020automatically,ma2020powertransformer} and paraphrasing \cite{van2019evaluating,xu2012paraphrasing} require meaning-preserving transformations, while political leaning \cite{prabhumoye2018style}, sentiment \cite{shen2017style,hu2017toward}, and topical transfer \cite{huang2020cycle} allow for a change in the underlying meaning. For instance, for a negative review, ``\textit{this was a bland dish},'' we can use a sentiment TST model to create a more positive ``\textit{this was a tasty dish},'' by swapping the word \textit{bland} with \textit{tasty.} Although the input's structure and attribute-independent content are preserved, the truth-conditional meaning is clearly altered. 

\begin{figure}
    \centering
    \includegraphics[width=0.98\columnwidth]{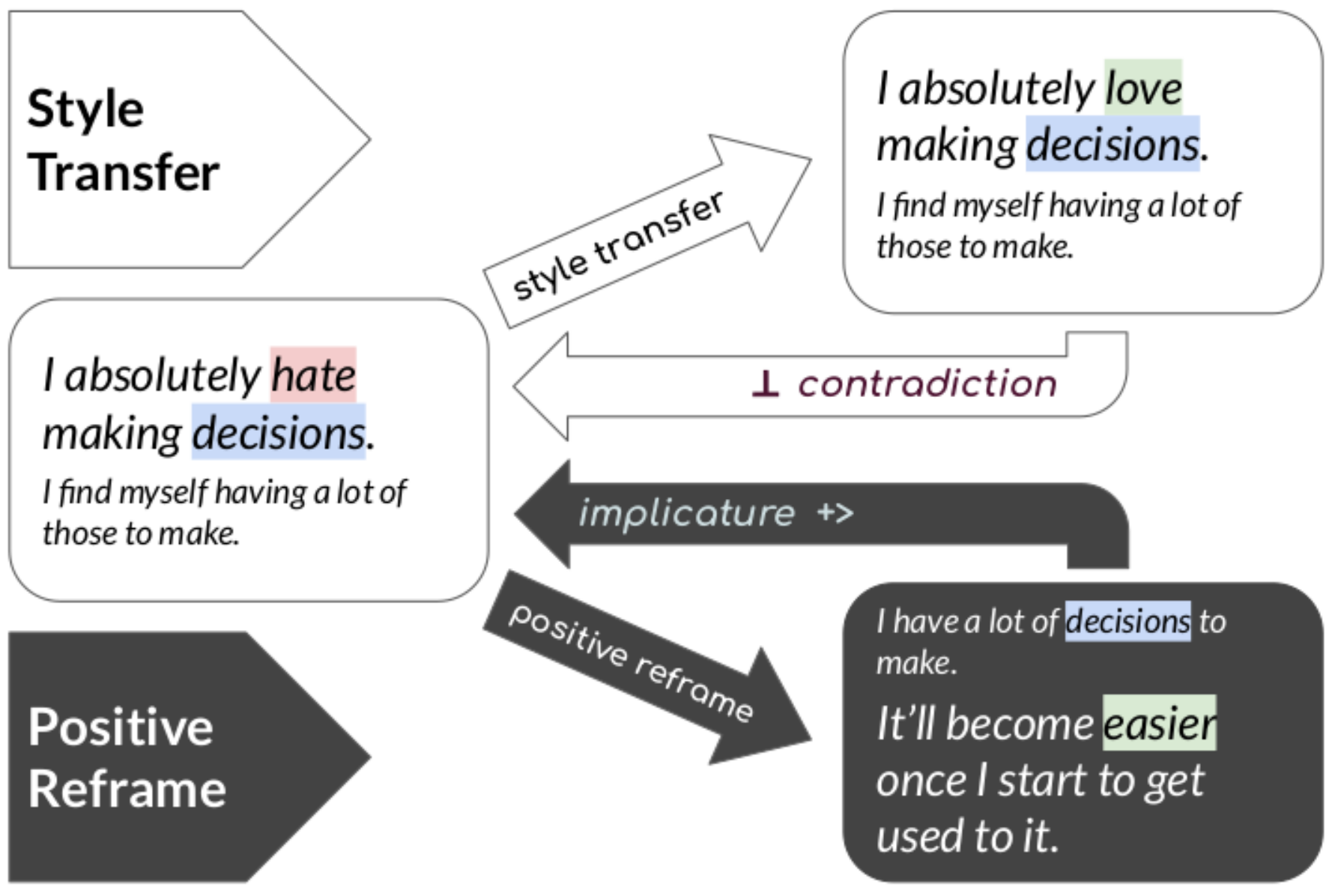}
    \caption{Positive reframing vs. negative-to-positive sentiment style transfer.
    }
    \label{fig:positive_reframing_example}
\end{figure}

In this work, we introduce a closely related task---positive reframing---that differs from sentiment TST in important ways. 
We effectively \textit{reframe} negative text by inducing a \textbf{complementary positive viewpoint} (e.g. \textit{glass-half-full}), which nevertheless supports the underlying content of the original sentence. The reframe should implicate rather than contradict the source (see Figure~\ref{fig:positive_reframing_example}), and the transformation should be motivated by 
theoretically justified strategies from from positive psychology (\citealt{harris2007integrating}; see Section~\ref{sec:framework}).

To use the example from before, we could reframe ``\textit{this was a bland dish}'' with the \textul{self-affirmation} strategy and say ``\textit{I've made dishes that are much tastier than this one}.'' This reframed one still communicates the author's original intention by conversationally implicating that the dish was unsatisfying \cite{grice1975logic}, but it shifts the focus away from the negative judgment and onto a positive and
self-affirming perspective. Numerous studies have shown the positive effects of this and other reframing strategies on well-being and cognitive performance \cite{martens2006combating,cohen2006reducing,good2003improving}, which motivate this work.

Our main contribution is the design and implementation of a new \textit{positive reframing} task. To facilitate research in this space, we introduce a parallel corpus of \numDatapoints{} reframed sentence pairs and \numStructuredAnnotations{} structured annotations for \numReframingStrategies{} theoretically-motivated re-write strategies. This is a significant contribution, especially since rich \emph{parallel} corpora are scarce in TST tasks. Some related datasets exist for politeness \cite{madaan2020politeness} and sentiment transfer \cite{shen2017style,he2016ups}, but they lack this parallel structure. With only unaligned corpora, researchers are limited to unsupervised training paradigms, which notoriously fail to disentangle style from content, and thus also fail to preserve meaning \cite{lample2018multiple}. Using our parallel corpus, we examine how current state-of-the-art neural models work for positive reframing. 
We find that, supervised transformer-based neural models appear capable of rewriting a negative text without contradicting the original premise of that text. However, these models still struggle to generate reasonable positive perspectives, suggesting that  our dataset will serve as a useful benchmark for understanding psychologically well-motivated strategies for augmenting text with positive perspectives.

\section{Related Work}

\subsection{Style-Transfer}
There is a longstanding interest in style transfer, starting with the early days schema-based systems \cite{mcdonald1985computational,hovy1987generating}, and then syntax-based \cite{zhu2010monolingual,xu2016optimizing} and phrase-based machine translation \cite{xu2012paraphrasing,wubben2012sentence}, into the age of end-to-end neural models. Recent works include supervised seq2seq tasks on parallel data \cite{rao2018dear,fu2018style} or pseudo-parallel data \cite{jin2019imat,zhang-etal-2020-parallel}, as well as unsupervised generative modeling on non-parallel data \cite{hu2017toward,shen2017style}, and semi-supervised techniques \cite{shang2019semi}. Other ideas include domain adaptation \cite{li-etal-2019-domain} or multi-task learning \cite{niu2018multi}, zero-shot translation \cite{korotkova2019grammatical}, unsupervised ``delete and generate'' approaches \cite{li2018delete,sudhakar2019transforming,malmi2020unsupervised,madaan2020politeness}, and reinforcement learning \cite{zhang2017sentence,wang2016text}.

Many existing datasets lack parallel structure, so the unsupervised setting is common in TST. Unfortunately, many of these methods still fail to disentangle style from content and adequately preserve the meaning of the original text \cite{lample2018multiple}. Autoencoders are particularly vulnerable to this shortcoming \cite{hu2017toward,zhao2018adversarially}, but some unsupervised machine translation techniques appear less vulnerable \cite{artetxe2018unsupervised,lample2018unsupervised}. 
In contrast, our positive reframing task requires source meaning-preservation and the introduction of \textit{new} content and \textit{new} perspectives, posing a unique challenge to unsupervised methods. We also provide a parallel corpus to train supervised models for this task.

\subsection{Language and Positive Psychology}
Positivity is contagious and can spread quickly across social networks \cite{coviello2014detecting,hatfield1993emotional}. Positive contagion in teams can reduce group conflict and improve group cooperation and even task performance \cite{barsade2002ripple}. 
Effective leaders also harness the power of positive reframing to promote company growth \cite{sy2013contagious, sy2005contagious,johnson2009you,masters1992use} and beneficially shape negotiations \cite{filipowicz2011understanding}, customer relations \cite{dietz2004service}, decision making \cite{gachter2009experimental,druckman2001using} and policy outcomes \cite{erisen2014affective}. At an individual level, people who express optimism and gratitude are less likely to have depressive symptoms \cite{lambert2012gratitude} and more likely to experience emotional and psychological well-being \cite{carver1999coping,watkins2008taking,scheier2001optimism}.

On the other hand, fake expressions of positivity are correlated with negative brain activity \cite{ekman1990duchenne} and may actually be more harmful than helpful \cite{fredrickson2000extracting,fredrickson2005positive,gross2013handbook,logel2009perils}. That is why in our task it is essential that any \textit{positively reframed} rephrased text remain true to the original premise of the source. In this way, our task is most similar to meaning-preserving transformations via parallel corpora from domains such as political argumentation \cite{chakrabarty2021entrust}, de-biasing \cite{pryzant2020automatically,ma2020powertransformer}, politeness \cite{madaan2020politeness}, and paraphrasing \cite{van2019evaluating,xu2012paraphrasing}.

\section{Positive Reframing Framework}
\label{sec:framework}

In this section, we present our psychologically-motivated taxonomy of positive reframing strategies. Instead of merely swapping antonyms for negative words or inserting unfounded positive language into a sentence, these strategies work to more fundamentally reconstruct the author's fixed, global, and ultimately harmful self-narratives, which are known in the literature as \textit{cognitive distortions} \cite{burns1981feeling,abramson2002cognitive,walton2020bad}. Cognitive distortions include many exaggerated or irrational self-focused thoughts \cite{nalabandian-ireland-2019-depressed}, such as dichotomous ``all-or-nothing'' thinking \cite{oshio2012all}, over-generalization \cite{muran1993cognitive}, and catastrophizing \cite{sullivan2001catastrophizing}. We can reconstruct these ideas using strategies from positive psychology \cite{harris2007integrating}. Each strategy is designed to promote a beneficial shift in perspective \textit{without distorting the underlying context} of the author's situation.

\paragraph{Growth Mindset} or, alternatively, the \textit{incremental theory of personality} \cite{yeager2014far,burnette2012buffering}, is the belief that one's skills and abilities are not immutable but can instead be changed and improved over time \cite{dweck2016having}; that one's willpower is an \textit{abundant} rather than limited or exhaustible resource \cite{job2010ego,Job2015ImplicitTA}; and that apparent setbacks like stress can be enhancing rather than debilitating \cite{crum2013rethinking}. Instead of saying ``\textit{I'm such a lazy procrastinator},'' a growth-mindset would say ``\textit{I'm determined to learn better time management}.'' This mindset has demonstrable benefits like improved performance on school tests \cite{good2003improving,blackwell2007implicit,dweck2019mindsets,yeager2014far}.

\paragraph{Impermanence} means understanding that negative experiences are finite and temporary, and that others have also experienced or even overcome similar forms of adversity. Someone might say ``\textit{since I failed this test, I must be too stupid for school}.'' An impermanence reframe could be ``\textit{This wasn't the test score I hoped for, but everyone slips up now and then.}'' This category is also related to those proposed by \citet{walton2020bad}: (1) focus on the ``possibility of improvement,'' (2) recognize ``specific, normal causes,'' and (3) understand ``you're not the only one.''

\paragraph{Neutralizing} involves removing or rewriting negative phrases and terms so they are more neutral \cite{pryzant2020automatically}. Someone might complain that ``\textit{Wendy’s customer service is terrible}.'' A neutralized reframe could be ``\textit{Wendy’s customer service could use some improvement}.'' 

\paragraph{Optimism} does not mean to negate or deny the negative aspects of a situation, but instead to shift the emphasis to the more positive aspects of the situation, including expectations for a bright future \cite{carver2010optimism}. For example, if there is a negative emphasis, like in the sentence, ``\textit{I've completely worked myself to the bone this week, burning the candle at both ends... TGIF},'' we can use optimism to shift the emphasis towards the positive as follows: ``\textit{It's been a long week, but now I can kick back, relax, and enjoy my favorite shows because it's the weekend}.'' 

\paragraph{Self-affirmation} means to assert a more holistic or expansive version of oneself by listing one's values, skills, and positive characteristics \cite{cohen2014psychology,silverman2013self}. Positive psychology gives many examples like love, courage, hope, gratitude, patience, forgiveness, creativity, and humor \cite{harris2007integrating}. Reflecting on these values can bolster one's sense of integrity (see Self-Affirmation Theory; \citealt{steele1988psychology}), can reduce depressive affect \cite{enright2000helping}, and can translate to increased performance on measurable tasks like exams \cite{martens2006combating,cohen2006reducing,sherman2009affirmed}.

\paragraph{Thankfulness} can also be described more broadly as an ``attitude of gratitude'' \cite{emmons2002gratitude}. Adding more positive words that convey thankfulness or gratitude (e.g. appreciate, glad that, thankful for). For example, we can reframe the rhetorical question ,``\textit{Is it sad that I don't wanna be at home and wish that work could call me in early}?'' by expressing gratitude for career: ``\textit{I am thankful that I have a job that makes me want to get out of bed everyday}.''

\begin{table*}
    \centering
        \resizebox{1\textwidth}{!}{%
        \begingroup
        \renewcommand{\arraystretch}{1.25} 
        \begin{tabular}{rllL{94mm}|rrr}
            \toprule
            \textbf{Label Distribution} & \textbf{Count} & \textbf{Label} & \textbf{Description} & \textbf{ICC} & \textbf{Gen} \\ \midrule
            \DrawPercentageBar{25.4}
            &2,120& Growth Mindset & Viewing a challenging event as an opportunity for the author specifically to grow or improve themselves. & 0.59 & 3.77\\\hline
            
            \DrawPercentageBar{19.5} &1,625& Impermanence & Saying bad things don't last forever, will get better soon, and/or that others have experienced similar struggles.& 0.60 & 4.03 \\\hline
            
            \DrawPercentageBar{36.1}&3,015& Neutralizing &Replacing a negative word with a neutral word. For example, ``This was a terrible day'' becomes ``This was a long day.'' & 0.32 & 3.53\\\hline
            
            \DrawPercentageBar{48.7} &4,069& Optimism & Focusing on things about the situation itself, in that moment, that are good (not just forecasting a better future).& 0.44& 3.89 \\\hline
            
            \DrawPercentageBar{10.1}&841& Self-affirmation &Talking about what strengths the author already has, or the values they admire, like love, courage, perseverance, etc. & 0.42 & 3.75\\ \hline
            
            \DrawPercentageBar{13.0} &1,085& Thankfulness & Expressing thankfulness or gratitude with key words like appreciate, glad that, thankful for, good thing, etc. & 0.68 & 3.95\\
            \bottomrule
        \end{tabular}
        \endgroup
        }
        \caption{\small{Summary statistics for \Data{}. (\textit{Left}) Distribution of non-exclusive labels across all \numDatapoints{} annotations shows a preference for \textit{optimism} and \textit{neutralizing} strategies. (\textit{Right}) The quality of annotations is shown by moderate Intra-class Correlation (ICC), with reasonable \textit{genuineness} (Gen) metrics for 100 randomly sampled datapoints.}}
        \label{fig:data_stats}
\end{table*}

\section{Data Collection}
We sourced all of our data from the Twitter API, filtering tweets according to the hashtag \texttt{\#stressed} due to a few reasons. Note that at the time of data collection and annotation, there were no publicly available datasets with annotated cognitive distortions, and the literature on distortion classification was still relatively unexplored \cite{simms2017detecting,shickel2020automatic}. We instead chose the simple keyword \texttt{\#stressed} to signal the anxiety, negative affect, and hopelessness that has been shown to accompany cognitive distortions by prior work \cite{sears2009think}.\footnote{We also considered \textsl{pet peeve}, \textsl{fml}, and other keywords but manual inspection revealed that these tweets were unlikely to contain cognitive distortions. In contrast, \textsl{stressed} hashtag provides a high precision data collection. We acknowledge this as a limitation and urge readers to keep this mind when interpreting our findings.
} Our decision to use Twitter was also motivated by the 280 character limit, which ensured that samples were short, focused expressions of relatively atomic ideas, as opposed to longer narrative-style texts from discussion platforms like Reddit's \texttt{r/rant}. 

Our filtered collection of negative texts comes from a collection of over 1 million \texttt{\#stressed} tweets written between 2012 and 2021, and it excludes any replies and retweets, any insubstantial tweets less than 30 characters, and any text containing a URL, which is often associated with spam \cite{zhang2012detecting,grier2010spam}. After we removed other hashtags or Twitter handles from the text, we used TextBlob \cite{loria2018textblob} to exclude any overtly positive texts with a non-negative sentiment score. Finally, to reduce any confounds between cognitive distortions and hate speech, and to make the human annotation task more agreeable for crowd-workers, we excluded examples that were flagged as offensive with over 80\% confidence according to HateSonar \cite{davidson2017automated}.

\subsection{Annotation}
\label{subsec:annotation}
We recruited crowdworkers to reframe 8,687 randomly-sampled texts with two workers assigned to each task, so we had two unique reframe annotations for every tweet. The annotators were encouraged to decide independently which reframing strategy to use, and they could combine multiple strategies in the same reframe. We simply asked annotators to record the strategies they selected. Additionally, they gave us, on a scale from 1-5, a score indicating how positive the original text was, and separately, how positive the text had become after they reframed it. Finally, we asked workers to mark advertisements, spam, or any text they felt they could not understand or effectively reframe. These examples were later removed from the corpus (see Appendix~\ref{appdx:data_quality_control} for details).

In total, \numWorkers{} workers participated in this task. Before they worked on the task, workers were asked to be familiar with our task by reading our provided reframing examples for each of the six strategies (Section~\ref{sec:framework}), along with detailed annotation instructions. Then they had to pass a qualification test to show they can recognize different strategies in different reframing examples, with at least 5 out of 6 multiple-choice questions answered correctly. 

We paid all annotators a fair wage above the federal minimum and both manually and programmatically inspected their work for quality (see Appendix~\ref{appdx:data_quality_control}). After removing any poor-quality data, we were left with \numDatapoints{} reframed sentences. The strategy label distribution is given on the left side of Table~\ref{fig:data_stats}, where a single reframe can have more than one strategy label. 

\subsection{Data Quality}
\label{subsec:data_quality}
To determine the reliability of the reframing strategy constructs, we randomly sampled 100 annotations from Section~\ref{subsec:annotation} and asked three annotators to consider both the original text and the reframed text, and then the annotators marked which of the six strategies were used in the given reframe. This allowed us to compute inter-annotator agreement scores for the strategy labels in Table~\ref{fig:data_stats}. We observe the Intra-class Correlation for one-way random effects between the three raters and find moderate inter-rater agreement across these attribute categories (min 0.32; max 68). We also asked this second round of annotators to evaluate the \textit{genuineness} of the reframes on a scale from 1-5. Our instructions explain that, with a more genuine reframe, it is more likely that someone in the original situation would say something similar. We find that, across all strategy labels, the average genuineness score is $\sim 4$ out of 5, so we know the data conforms reasonably well to our task instructions.

\section{Positive Reframing}
\label{sec:generation}

\begin{table*}[h!]
\resizebox{\textwidth}{!}{%
\def\arraystretch{1.15}
\begin{tabular}{cclccccccc|ccc}\toprule
&& \multicolumn{8}{c}{\textbf{Automatic Evaluation}} & \multicolumn{3}{c}{\textbf{Human Evaluation}}\\
 &&\textbf{Model} & \Meaning{R-1} & \Meaning{R-2} & \Meaning{R-L} & \Meaning{BLEU} & \Meaning{BScore} & \Positivity{$\Delta$ TB} & \Fluency{Avg. Len} & \Meaning{Meaning} & \Positivity{Positivity} & \Fluency{Fluency}\\ \midrule
 \multicolumn{2}{c}{Retrieval}& Random& 9.6 & \textbf{3.6} & 8.4 & 0.17 & 84.8 & \textbf{0.36} & \textbf{20.0} & 2.79 & 3.03 & 3.60\\
 && SBERT& \textbf{15.2} & 1.9 & \textbf{12.8} & \textbf{1.47} & \textbf{87.6} & \textbf{0.36} &17.7 & \textbf{3.45} & \textbf{3.97} & \textbf{4.16}\\[1.5mm] \hline
 
  \multicolumn{2}{c}{Few-shot}& GPT-3 & 18.3 & \textbf{3.4} & 15.5 & 2.9 & \textbf{88.2} & \textbf{0.44} & 17.3 & \textbf{3.73} & \best{\textbf{4.17}} & \best{\textbf{4.27}}\\
 && GPT-Neo & \textbf{18.7} & \textbf{3.4} & \textbf{16.0} & \textbf{3.0} & \textbf{88.2} & 0.40 & \textbf{17.6} & 3.69 & 4.16 & 4.21\\[1.5mm]
 \hline
 
 \rule{0pt}{1.5\normalbaselineskip}\parbox[t]{2mm}{\multirow{7}{*}{\rotatebox[origin=c]{90}{Unconstrained}}}& \parbox[t]{2mm}{\multirow{7}{*}{\rotatebox[origin=c]{90}{$p(t|s)$}}}&GPT& 13.3 & 1.8 & 11.3 & 1.1 & 86.4 & 0.37 &21.1 & 3.55 & 3.91 & 4.08\\
 &&GPT-2 No-pretrain & 13.2 & 1.3 & 11.4 & 0.66 & \best{\textbf{89.6}} & 0.37& 16.9 & 3.11 & 3.66 & 3.96 \\
 &&GPT-2 & 20.9 & 4.6 & 17.7 & 4.2 & 88.5 & 0.35 & 20.0 & 3.58 & 4.01 & \textbf{4.18}\\
 &&Seq2Seq-LSTM & 15.7 & 1.4 & 12.4 & 0.73 & 85.6 & \best{\textbf{0.49}} & 25.8 & 3.33 & 4.15 & 4.10\\
 &&CopyNMT & 20.8 & 5.0 & 18.0 & 4.0 & 85.7 & 0.32& 16.1 & 3.57 & 3.69 & 3.91\\
 &&T5 & 27.4 & 9.8 & 23.8 & 8.7 & 88.7& 0.38 & \best{\textbf{35.3}} & 4.09 & 3.79 & 4.06\\
 &&BART & \textbf{27.7} & \textbf{10.8} & \textbf{24.3} & \best{\textbf{10.3}} & 89.3 & 0.23 & 24.4 & \textbf{4.13} & \textbf{3.81} & 4.15\\[1.5mm]
 
 \midrule
 \parbox[t]{4mm}{\centering \multirow{4}{*}{\rotatebox[origin=c]{90}{\phantom{abc}\makecell{Predict}}}} &\parbox[t]{4mm}{ \centering \multirow{4}{*}{\rotatebox[origin=c]{90}{\phantom{abc}$p(t, \boldsymbol{\psi}_t | s)$}}} & & & & & & & & & \\[-2.5mm]
 &&T5 & \textbf{27.5} & \textbf{10.5} & 24.0 & \textbf{11.0} & 89.0 & 0.23 & \textbf{25.1} & \textbf{4.10} & 3.64 & \textbf{4.11} \\
 &&BART & 27.3 & 10.2 & \textbf{24.1} & 9.85 & \textbf{89.4} & \textbf{0.32} & 23.4 & 4.09 & \textbf{3.95} & \textbf{4.11}  \\ 
 & & & & & & & & & \\[-2.5mm] 
 
 \midrule
 \parbox[t]{4mm}{\multirow{4}{*}{\rotatebox[origin=c]{90}{\makecell{\phantom{abc}Control}}}} &\parbox[t]{4mm}{\multirow{4}{*}{\rotatebox[origin=c]{90}{\phantom{abc}$p(t|s, \boldsymbol{\psi}_t)$}}}  & & & & & & & & & \\[-2.5mm]
 &&T5 & 27.7 & 10.0 & 23.9 & 8.8 & 88.8 & \textbf{0.36} & \textbf{35.0} & 4.11 & 3.89 & 4.07\\
 &&BART & \best{\textbf{28.8}} & \best{\textbf{10.9}} & \best{\textbf{25.1}} & \textbf{10.1} & \best{\textbf{89.6}} & 0.27 & 24.7 & \best{\textbf{4.23}} & \textbf{4.07} & \best{\textbf{4.27}} \\ 
  & & & & & & & & & \\[-2.5mm] \hline
  
 \midrule 
 
 \rule{0pt}{\normalbaselineskip}&& \textit{Human} & \textit{100} & \textit{100} & \textit{100} & \textit{100} & \textit{100} & \textit{0.35} & \textit{17.4} & \textit{3.80} & \textit{3.82} & \textit{4.18} \\[1.5mm]
\bottomrule 
\end{tabular}
}
\caption{\small{\textbf{Positive reframing results} 
meausred by \Meaning{Meaning} including  \Meaning{ROUGE-1 (R-1), ROUGE-1 (R-2), ROUGE-L (R-L)}, \Meaning{BLEU}, \Meaning{BERTScore (BScore)}, \Positivity{Positivity} via
\Positivity{$\Delta$ TextBlob ($\Delta$ TB)} and \Fluency{Fluency}. State-of-the-art models can generate meaning-preserving reframes in the unconstrained setting $p(t|s)$ and strategy-predictive setting $p(t, \boldsymbol{\psi}_t | s)$ as well as when we condition the generation to use the reframing strategy from the ground truth $p(t|s, \boldsymbol{\psi}_t)$. The best in-category performance is \textbf{bolded}; best overall performance is \best{\textbf{highlighted}}.}}
\label{tab:generation_results}
\end{table*}

With \Data{}, we then examine how generative models work to automatically suggest 
a negatively-oriented self-narrative with a more positive shift in perspective without distorting any of the underlying meaning of that text. To do so will make use of encoder-decoder or conditional language models, as well as the six positive psychology strategies outlined in Section~\ref{sec:framework}.

\subsection{Task Formulation}
Let $(s, t, \boldsymbol{\psi}_t)$ be a single annotation tuple in \Data{} for original source text $s$ and positive reframe target $t$, which uses positive psychology strategies given by the multi-hot encoded vector $\boldsymbol{\psi}_t$. In the Positive Reframing task, our goal is to encode $s$ and, at decoding time, produce $t$ which makes use of $\boldsymbol{\psi}_t$ strategies and preserves the underlying meaning of $s$. Therefore, we formulate the problem as conditional generation and, during training, we maximize the standard language modeling objective
\begin{equation*}
    \label{eq:objective}
    \frac{1}{N} \sum_{i=0}^N \log{p(g_i | g_{0:i-1})}
\end{equation*}
over the string 
\begin{align*}
    \boldsymbol{g} &= \{s, \boldsymbol{\psi}_t, t\}\\
    &= \{\texttt{<BOS>}, s_1, s_2, ..., s_n, \\
    & \phantom{abc} \texttt{<STRG>}, \psi_{\texttt{grow}}, \psi_{\texttt{imp}}, ..., \psi_{\texttt{thank}}, \\
    & \phantom{abc} \texttt{<REFR>}, t_{1}, t_{2}, ..., t_{m}, \texttt{<EOS>}\}
\end{align*}
where $g_i$ is the $i$th token in the string of length $N$, which contains the start token \texttt{<BOS>}, the tokenized source $s_{1:n}$, the tokenized reframe target $t_{1:m}$, and the binary tokens $\psi_{\texttt{grow}}, \psi_{\texttt{imp}}, ...$ indicating whether a particular strategy (e.g. \texttt{grow}th mindset) was used in reframe $t$. 

At decoding time, we consider three settings: \textit{Unconstrained} generation $p(t|s)$, \textit{Controlled} generation $p(t|s, \boldsymbol{\psi}_t)$, and a strategy \textit{Prediction} form of generation $p(t, \boldsymbol{\psi}_t|s)$. Unlike in the \textit{Unconstrained} setting, the \textit{Controlled} generation is conditioned on the desired strategies $\boldsymbol{\psi}_t$. In the \textit{Prediction} setting, the model will concurrently predict the strategies it used to generate its own reframe. 

Note that, we introduce three different model settings here to capture how positive reframing assistance might be used by people in the real world. 
Specifically, 
the \textit{Unconstrained} setting models reframing text directly without being aware of any specific strategy to use. 
The \textit{Prediction} setting  
extends the unconstrained mode, i.e., produce the reframed text and also output the reframing strategies used in the reframing process spontaneously.  
The \textit{Controlled} setting 
simulates the scenario of producing a reframed text with the help of concrete positive reframing strategies. 

\subsection{Experimental Setup}
For ground truth training, development, and testing, we randomly partition the annotations using an 8:1:1 ratio, with 6,679 train, 835 development and 835 test data. We fine-tune the GPT and GPT-2 language models \cite{radford2019language} as well as two Seq2Seq neural machine translation models -- LSTM \cite{hochreiter1997long} and CopyNMT \cite{see2017get} -- and finally, two encoder-decoder models, BART \cite{lewis2020bart} and T5 \cite{raffel2020exploring}. For all models, we use greedy decoding. As an ablation in the \textit{Unconstrained} setting, we also test a \textit{No-pretrain} condition for GPT-2 in which we randomly initialize the model parameters before fine-tuning. 

\textbf{Retrieval:} We test two simple retrieval systems: \textit{Random} retrieval of a reframed sentence from the training set, and SBERT \cite{reimers-2019-sentence-bert} retrieval, which finds the most similar $t$ in train by cosine similarity and retrieves one of the corresponding ground-truth $r$ from the training set.

\textbf{Few-shot Learning:} \citet{brown2020language} shows the few-shot capabilities of language models and especially larger models like GPT-3. We evaluate few-shot abilities of both GPT-3 and its open-source implementation, GPT-Neo \cite{gpt-neo} using $k=5$ exemplars (See Appendix~\ref{appdx:few_shot}).

\subsection{Evaluation} Following other style transfer work with a parallel corpus \cite{jhamtani2017shakespearizing,xu2012paraphrasing}, we evaluate our models for semantic similarity with the ground truth using the {BLEU}~\cite{papineni2002bleu}, {ROUGE} \cite{lin2004rouge}, and {BERTScore}~\cite{zhang2019bertscore}. Since there are two ground truth annotations per tweet, we take the maximum of the two scores and report the average across these maxima. We also report {$\Delta TextBlob$} or the average change in sentiment score according to TextBlob \cite{loria2018textblob}. 
Finally, we conduct human evaluation in which \numHumanEvals{} items are distributed to 3 raters who score the reframed sentences for three criteria, each on a scale from 1 to 5. The criteria include {\textit{Meaning Preservation}} \cite{shang2019semi}, our task-specific objective, as well as the {\textit{Positivity}} and {\textit{Fluency}} of the generated text, following the sentiment style transfer literature \cite{luo2019towards} 

\subsection{Results}
\label{subsec:results}
\paragraph{Automatic Evaluation} Across these metrics (Table~\ref{tab:generation_results}, \textit{left}) in the unconstrained generation setting, the BART model provided the highest quality of positive reframes, while GPT provided the worst quality with results similar to the \textit{No-pretrain} version of GPT-2. The pre-trained version of GPT-2 was trained on English web text, while GPT was trained on works of fiction, so it appears that pre-training decisions can affect performance. 

We tested the two best-performing models, T-5 and BART, on the controlled generation and strategy-prediction settings as well and found that the both models performed reasonably. Overall, controlled generation boosts performance, since the model can target the gold standard's strategies, but these improvements are only slight (see the \textit{Controlled} part in  Table~\ref{tab:generation_results}). This warrants further investigation: in Section~\ref{sec:classification}, we explore models' ability to identify the underlying strategies given an existing reframe to understand whether models can make sense of these underlying constructs.

\begin{table*}[t]
\resizebox{\textwidth}{!}{
\begin{tabular}{L{65mm}L{20mm}L{130mm}}
\toprule
\textbf{Original Text} & \textbf{Strategy}  & \textbf{Positive Reframe}\\ \midrule
(a) Could someone just give me like \$1000? It would change my life, stressing about rent, bills and food money is just the worst. & growth & {I need to learn how to manage my money better so that I don't have to stress about rent, bills and food money.} \newline \textit{Time to focus on making a budget, so I don't have to stress about rent, bills, and food money.} \\ \midrule
(b) I just went back to school today And I'm already stressed cause we have MIDTERMS NEXT WEEK !!! AND THIS WEEKEND -- JAM PACKED \text{:-(}   & impermanence  & {I just went back to school today and I'm already stressed because we have midterms next week and this weekend -- JAM PACKED} \newline \textit{Well, just think by the end of next week we'll be all done with the midterms!} \\ \midrule
(c) talacre beach really need to sort there checking in routeen caravans are never ready 30mins later and still sat in the car waiting .. & neutralizing & The caravans at Talacre beach need to sort out their check-in process, so the kids don't get stuck in the car. \newline \textit{Talacre beach's check in process could do with some improvement to reduce the wait time for customers.} \\ \midrule
(d) So glad that tomorrow is Friday. This has seriously been the longest week of my life! tired & optimism & I'm glad that tomorrow is Friday. This week has been long, but I'm looking forward to the weekend. \newline \textit{I'm glad the weekend is coming up, so I can rest.} \\ \midrule
(e) Sometimes I get these impulses to just throw a tantrum. Like throw/break things, cry and scream. & self-affirmation & Sometimes I get these impulses to just throw a tantrum. Like throw/break things, cry and scream. But I'm strong, and I know I can handle it. \newline \textit{It's normal for to feel overwhelmed sometimes but I know I am strong to handle and go through it.} \\ \midrule
(f) The fact that my mom's been calling me every night to calm me down from school says so much. I've needed it these past few days & thankfulness & I'm really lucky to have such a caring mum who is willing to call me every night when I'm having a hard time. \newline \textit{My mom has been calling me every night to calm me down from school. I've needed it these past few days. I'm thankful for her.}
\\ \bottomrule
\end{tabular}}
\caption{\small{\textbf{Example reframes.} We examine one original text for each of the strategies in our dataset along with its BART-generated positive reframe (first row of each block) and its ground truth reframe (\textit{italics} in second row of each block). Even when the generations differ from the ground truth, the model's reframes are largely successful at shifting the perspective while still maintaining the underlying meaning of the original text.}
} \label{tab:posts_generation_examples}
\end{table*}

Unsurprisingly, all supervised models outperformed our simple retrieval baselines. Most interestingly, few-shot GPT-3 and GPT-Neo also \textbf{could not} match the supervised models in terms of overlap with the ground truth (\Meaning{ROUGE}, \Meaning{BLEU}, \Meaning{BERTScore}), but they still achieved a comparable positive shift in sentiment (\Positivity{$\Delta$ TextBlob}). 

\paragraph{Human Evaluation} Human judgments both support and elaborate on the automatic evaluation findings. For our best performing BART and T-5 models, the average scores are very high, even surpassing the quality of the \textit{Human} gold standard in all of the unconstrained, predictive, and controlled settings. These systems most effectively induce a natural-sounding positive reframe while also \textit{preserving the meaning} of the original text. This is critical: controlled BART model scored 4.07 in \Positivity{Positivity} and 4.27  in \Fluency{Fluency} while also achieving the winning \Meaning{Meaning} preservation score. 

In contrast with BART, the few-shot systems fail to preserve the meaning of the original sentence, despite their ability to articulately induce a more positive sentiment (\Positivity{Positivity} scores up to 4.17; \Fluency{Fluency} scores up to 4.27). Meaning preservation is absolutely critical for this task. From these results, we can conclude that, at the present time, supervised learning may be the most viable option for achieving reliable positive reframing results. \Data{}{} will facilitate ongoing efforts in this direction.

\paragraph{Qualitative Investigation} Table~\ref{tab:posts_generation_examples} shows example reframes generated by our best controlled BART model, with one example for each strategy (for a similar comparison between \textit{models}, see Table~\ref{tab:generation_model_comparison} in Appendix~\ref{appdx:example_reframes}). We see that, even without explicit lexical overlap between the generation and ground truth, the model reframes can still shift the cognitive distortions and negative outlook to a more positive perspective. In each of these examples, the model does so without losing the underlying meaning of the original text. Transformer-based models appear to be capable of solving our task with reasonable success. However, success can be highly variable (as evidenced by Table~\ref{tab:generation_model_comparison}), so there is still room for significant improvement.

\subsection{Error Analysis}\label{sec:error}
We manually go through 100 randomly sampled model generations by our best controlled BART model, and summarize the main error classes here.
We manually investigated 100 randomly sampled model generations by our best controlled BART model, and summarize the four largest error classes here. First, 26\% of generations contained \textbf{(1) insubstantial changes}. These were especially prominent in the \textit{neutralizing} strategy where the model would swap only a few negative words, like changing the phrase ``\textit{I hate it}'' to ``\textit{I don't like it.}'' On the other hand, some reframed generations were so drastically modified they contained \textbf{(2) contradictions to the premise} (9\% of instances). For example, "\textit{Feel like crying, this math class is impossible to pass}" was transformed into "\textit{This math class is hard, but I know I can pass it}" -- a failure of meaning preservation. More concerningly, the system can generate \textbf{(3) self-contradictions} (6\%) like the phrase, "\textit{I don't like opening up to people, but I'm glad I have the courage to do it.}" Finally, like many other NLG systems, our system can produce \textbf{(4) hallucinations} (2\%) with unmotivated perspectives, like mentioning a \textit{good night sleep} when the original post was about nosebleeds in the bath. 

\begin{table}[t]
\centering
\resizebox{7cm}{!}{
\begin{tabular}{@{}lcccc@{}}
\toprule
\textbf{Strategy}        & \textbf{BERT} & \textbf{RoBERTA} & \textbf{XLNet}   & \textbf{Support} \\ \midrule
Thankfulness     & 0.71      & 0.69   & 0.71 & 109     \\
Neutralizing     & 0.59      & 0.60   & 0.49 & 302     \\
Optimism         & 0.72      & 0.71   & 0.72 & 400     \\
Impermanence     & 0.55      & 0.55   & 0.54 & 157     \\
Growth           & 0.61      & 0.63   & 0.67 & 221     \\
Self Affirmation & 0.43      & 0.44   & 0.39 & 76      \\ \bottomrule
\end{tabular}}
\caption{Strategy classification F1 scores
} 
\label{tab:classification_results}
\end{table}
\subsection{Frame Strategy Classification}
\label{sec:classification}

In Section~\ref{subsec:results}, we observed only slight performance gains when conditioning the generation based on the ground-truth reframing strategy (\textit{Control} section in Table~\ref{tab:generation_results}). For this reason, we
take a closer look at whether models can reliably understand and classify the reframe strategies underlying a given source-reframe text pair. 
We formulate this problem as a multi-label multi-class classification task over sentence pairs $(s,t)$. Given both the source text and positive reframe target in the annotation tuple  $(s, t)$ from \Data{}, we
predict the multi-hot encoded strategy vector $\boldsymbol{\psi}_t = [s_{\texttt{grow}}; s_{\texttt{imp}}; ...; s_{\texttt{thank}}]$ using transformer models. 
We experiment with a set of state-of-the-art classifiers, including BERT \cite{devlin2019bert}, RoBERTA \cite{liu2019roberta}, and XLNet \cite{yang2019xlnet}.

As shown in Table~\ref{tab:classification_results}, all of the classification models can learn to recognize the thankfulness, optimism, and growth mindset strategies with moderate reliability $(F_1>0.60)$. Although XLNet model cannot identify the neutralizing strategy very well, BERT and RoBERTa models can achieve an F1 score of around 0.6. The impermanence and self-affirmation strategies appear more challenging for all three models to identify. 
Overall, the results here show that this task is tractable: reframe strategies are \textbf{learnable} by various classification models. This further supports the reliability of our Positive Psychology framework, confirming what we found with human reliability metrics in Section~\ref{subsec:data_quality}.
Although we mainly treat this frame strategy classification as a robustness check and deep dive into the role of framing strategies, this task can also be a novel NLP or computational social science application on its own, i.e., determining the positive reframing relation between a pair of sentences. 

\section{Discussion and Conclusion}
This work introduces a new and challenging NLG task called \textit{positive reframing}. The objective is to construct a more positive outlook as a way of rephrasing a negative source text such that the meaning of that source is preserved. Our parallel dataset, \Data{}{}, will serve as a benchmark that will enable sustained work on this task. We experiment with many of the leading style-transfer models and show that these models can learn to shift from a negative to a more positive perspective using a combination of strategies from positive psychology. Importantly, the best models are fluent and effective reframing systems that can learn to largely preserve the meaning of the original text, even under a perspective shift.   However, these models still struggle to generate reasonable positive perspectives, and  even the best models are still prone to errors. We discuss four key error classes: insubstantial changes, contradictions to the premise, self-contradictions, and hallucinations, 
as shown in Error Analyses in Section \ref{sec:error}. Overall, this suggests that  our dataset can serve as a useful benchmark for understanding well-motivated positive reframing strategies and equipping natural language generation systems with positive perspectives.

Future work can dive deeper into these issues by enforcing a stronger level of semantic equivalence between the generation and the source text \cite{nie2019simple}. Even with semantic equivalence constraints, it would be necessary to also allow for the injection of new positive perspectives. Methods ranging from guided sequence generation \cite{krause2020gedi} or semantic attention-guided decoding \cite{nie2019simple} to pragmatic reconstruction \cite{shen2019pragmatically} and persona consistency \cite{kim2020will} may all be applicable in follow-up studies.

\section*{Acknowledgements} The authors would like to thank reviewers for their helpful insights and feedback. CZ is supported by the NSF Graduate Research Fellowship under Grant No. DGE-2039655 and DY is supported by the Microsoft Research Faculty Fellowship. This work is 
funded in part by a grant from Amazon.

\section*{Ethics}
\textbf{Annotation.} We followed the guidelines for ethical annotation practices and crowdsourcing that are outlined in \cite{sheehan2018crowdsourcing}, including paying workers a fair wage above the federal minimum. If workers contacted us with any questions or concerns, we responded promptly to them within 24 hours.  In the task interface, in the header, we warned annotators that the content might be upsetting, and we gave the following recommendation: ``\textit{if any point you do not feel comfortable, please feel free to skip the HIT or take a break.}''. 

\textbf{Deployment.} Although this data is designed for pro-social outcomes (i.e. increasing positivity in text), there may be unexpected use-cases for this data, such as obfuscating impolite or even hateful data to avoid detection \cite{elsherief-etal-2021-latent}. The parallel structure of the data means it is also possible to invert the direction of the seq2seq task to introduce more negative or pessimistic perspectives into a positive source. This is not a particularly new risk, since sentiment style transfer can accomplish a similar outcome in this direction. Still, we will require interested parties to sign a data-use agreement that encourages only ethical uses of \Data{}.

\bibliographystyle{acl_natbib}

\begin{thebibliography}{115}
\expandafter\ifx\csname natexlab\endcsname\relax\def\natexlab#1{#1}\fi

\bibitem[{Abramson et~al.(2002)Abramson, Alloy, Hankin, Haeffel, MacCoon, and
  Gibb}]{abramson2002cognitive}
Lyn~Y Abramson, Lauren~B Alloy, Benjamin~L Hankin, Gerald~J Haeffel, Donal~G
  MacCoon, and Brandon~E Gibb. 2002.
\newblock Cognitive vulnerability-stress models of depression in a
  self-regulatory and psychobiological context.

\bibitem[{Artetxe et~al.(2018)Artetxe, Labaka, and
  Agirre}]{artetxe2018unsupervised}
Mikel Artetxe, Gorka Labaka, and Eneko Agirre. 2018.
\newblock \href {https://doi.org/10.18653/v1/D18-1399} {Unsupervised
  statistical machine translation}.
\newblock In \emph{Proceedings of the 2018 Conference on Empirical Methods in
  Natural Language Processing}, pages 3632--3642, Brussels, Belgium.
  Association for Computational Linguistics.

\bibitem[{Baldini~Soares et~al.(2019)Baldini~Soares, FitzGerald, Ling, and
  Kwiatkowski}]{soares2019matching}
Livio Baldini~Soares, Nicholas FitzGerald, Jeffrey Ling, and Tom Kwiatkowski.
  2019.
\newblock \href {https://doi.org/10.18653/v1/P19-1279} {Matching the blanks:
  Distributional similarity for relation learning}.
\newblock In \emph{Proceedings of the 57th Annual Meeting of the Association
  for Computational Linguistics}, pages 2895--2905, Florence, Italy.
  Association for Computational Linguistics.

\bibitem[{Barsade(2002)}]{barsade2002ripple}
Sigal~G Barsade. 2002.
\newblock The ripple effect: Emotional contagion and its influence on group
  behavior.
\newblock \emph{Administrative science quarterly}, 47(4):644--675.

\bibitem[{Black et~al.(2021)Black, Gao, Wang, Leahy, and Biderman}]{gpt-neo}
Sid Black, Leo Gao, Phil Wang, Connor Leahy, and Stella Biderman. 2021.
\newblock \href {http://github.com/eleutherai/gpt-neo} {{GPT-Neo}: Large scale
  autoregressive language modeling with mesh-tensorflow}.

\bibitem[{Blackwell et~al.(2007)Blackwell, Trzesniewski, and
  Dweck}]{blackwell2007implicit}
Lisa~S Blackwell, Kali~H Trzesniewski, and Carol~Sorich Dweck. 2007.
\newblock Implicit theories of intelligence predict achievement across an
  adolescent transition: A longitudinal study and an intervention.
\newblock \emph{Child development}, 78(1):246--263.

\bibitem[{Brown et~al.(2020)Brown, Mann, Ryder, Subbiah, Kaplan, Dhariwal,
  Neelakantan, Shyam, Sastry, Askell, Agarwal, Herbert{-}Voss, Krueger,
  Henighan, Child, Ramesh, Ziegler, Wu, Winter, Hesse, Chen, Sigler, Litwin,
  Gray, Chess, Clark, Berner, McCandlish, Radford, Sutskever, and
  Amodei}]{brown2020language}
Tom~B. Brown, Benjamin Mann, Nick Ryder, Melanie Subbiah, Jared Kaplan,
  Prafulla Dhariwal, Arvind Neelakantan, Pranav Shyam, Girish Sastry, Amanda
  Askell, Sandhini Agarwal, Ariel Herbert{-}Voss, Gretchen Krueger, Tom
  Henighan, Rewon Child, Aditya Ramesh, Daniel~M. Ziegler, Jeffrey Wu, Clemens
  Winter, Christopher Hesse, Mark Chen, Eric Sigler, Mateusz Litwin, Scott
  Gray, Benjamin Chess, Jack Clark, Christopher Berner, Sam McCandlish, Alec
  Radford, Ilya Sutskever, and Dario Amodei. 2020.
\newblock \href
  {https://proceedings.neurips.cc/paper/2020/hash/1457c0d6bfcb4967418bfb8ac142f64a-Abstract.html}
  {Language models are few-shot learners}.
\newblock In \emph{Advances in Neural Information Processing Systems 33: Annual
  Conference on Neural Information Processing Systems 2020, NeurIPS 2020,
  December 6-12, 2020, virtual}.

\bibitem[{Burnette and Finkel(2012)}]{burnette2012buffering}
Jeni~L Burnette and Eli~J Finkel. 2012.
\newblock Buffering against weight gain following dieting setbacks: An implicit
  theory intervention.
\newblock \emph{Journal of Experimental Social Psychology}, 48(3):721--725.

\bibitem[{Burns(1981)}]{burns1981feeling}
David~D Burns. 1981.
\newblock \emph{Feeling good}.
\newblock Signet Book.

\bibitem[{Carver et~al.(1999)Carver, Pozo, Harris, Noriega, Scheier, Robinson,
  Ketcham, Moffat~Jr, and Clark}]{carver1999coping}
Charles~S Carver, Christina Pozo, Suzanne~D Harris, Victoria Noriega, Michael~F
  Scheier, David~S Robinson, Alfred~S Ketcham, Frederick~L Moffat~Jr, and
  Kimberley~C Clark. 1999.
\newblock How coping mediates the effect of optimism on distress: a study of
  women with early stage breast cancer.

\bibitem[{Carver et~al.(2010)Carver, Scheier, and
  Segerstrom}]{carver2010optimism}
Charles~S Carver, Michael~F Scheier, and Suzanne~C Segerstrom. 2010.
\newblock Optimism.
\newblock \emph{Clinical psychology review}, 30(7):879--889.

\bibitem[{Chakrabarty et~al.(2021)Chakrabarty, Hidey, and
  Muresan}]{chakrabarty2021entrust}
Tuhin Chakrabarty, Christopher Hidey, and Smaranda Muresan. 2021.
\newblock \href {https://doi.org/10.18653/v1/2021.naacl-main.394} {{ENTRUST}:
  Argument reframing with language models and entailment}.
\newblock In \emph{Proceedings of the 2021 Conference of the North American
  Chapter of the Association for Computational Linguistics: Human Language
  Technologies}, pages 4958--4971, Online. Association for Computational
  Linguistics.

\bibitem[{Cohen et~al.(2006)Cohen, Garcia, Apfel, and
  Master}]{cohen2006reducing}
Geoffrey~L Cohen, Julio Garcia, Nancy Apfel, and Allison Master. 2006.
\newblock Reducing the racial achievement gap: A social-psychological
  intervention.
\newblock \emph{science}, 313(5791):1307--1310.

\bibitem[{Cohen and Sherman(2014)}]{cohen2014psychology}
Geoffrey~L Cohen and David~K Sherman. 2014.
\newblock The psychology of change: Self-affirmation and social psychological
  intervention.
\newblock \emph{Annual review of psychology}, 65:333--371.

\bibitem[{Coviello et~al.(2014)Coviello, Sohn, Kramer, Marlow, Franceschetti,
  Christakis, and Fowler}]{coviello2014detecting}
Lorenzo Coviello, Yunkyu Sohn, Adam~DI Kramer, Cameron Marlow, Massimo
  Franceschetti, Nicholas~A Christakis, and James~H Fowler. 2014.
\newblock Detecting emotional contagion in massive social networks.
\newblock \emph{PloS one}, 9(3):e90315.

\bibitem[{Crum et~al.(2013)Crum, Salovey, and Achor}]{crum2013rethinking}
Alia~J Crum, Peter Salovey, and Shawn Achor. 2013.
\newblock Rethinking stress: the role of mindsets in determining the stress
  response.
\newblock \emph{Journal of personality and social psychology}, 104(4):716.

\bibitem[{Davidson et~al.(2017)Davidson, Warmsley, Macy, and
  Weber}]{davidson2017automated}
Thomas Davidson, Dana Warmsley, Michael Macy, and Ingmar Weber. 2017.
\newblock \href {https://arxiv.org/abs/1703.04009} {Automated hate speech
  detection and the problem of offensive language}.
\newblock \emph{ArXiv preprint}, abs/1703.04009.

\bibitem[{den Bercken et~al.(2019)den Bercken, Sips, and
  Lofi}]{van2019evaluating}
Laurens~Van den Bercken, Robert{-}Jan Sips, and Christoph Lofi. 2019.
\newblock \href {https://doi.org/10.1145/3308558.3313630} {Evaluating neural
  text simplification in the medical domain}.
\newblock In \emph{The World Wide Web Conference, {WWW} 2019, San Francisco,
  CA, USA, May 13-17, 2019}, pages 3286--3292. {ACM}.

\bibitem[{Devlin et~al.(2019)Devlin, Chang, Lee, and
  Toutanova}]{devlin2019bert}
Jacob Devlin, Ming-Wei Chang, Kenton Lee, and Kristina Toutanova. 2019.
\newblock \href {https://doi.org/10.18653/v1/N19-1423} {{BERT}: Pre-training of
  deep bidirectional transformers for language understanding}.
\newblock In \emph{Proceedings of the 2019 Conference of the North {A}merican
  Chapter of the Association for Computational Linguistics: Human Language
  Technologies, Volume 1 (Long and Short Papers)}, pages 4171--4186,
  Minneapolis, Minnesota. Association for Computational Linguistics.

\bibitem[{Dietz et~al.(2004)Dietz, Pugh, and Wiley}]{dietz2004service}
Joerg Dietz, S~Douglas Pugh, and Jack~W Wiley. 2004.
\newblock Service climate effects on customer attitudes: An examination of
  boundary conditions.
\newblock \emph{Academy of management journal}, 47(1):81--92.

\bibitem[{Druckman(2001)}]{druckman2001using}
James~N Druckman. 2001.
\newblock Using credible advice to overcome framing effects.
\newblock \emph{Journal of Law, Economics, and Organization}, 17(1):62--82.

\bibitem[{Dweck(2016)}]{dweck2016having}
Carol Dweck. 2016.
\newblock What having a “growth mindset” actually means.
\newblock \emph{Harvard Business Review}, 13:213--226.

\bibitem[{Dweck and Yeager(2019)}]{dweck2019mindsets}
Carol~S Dweck and David~S Yeager. 2019.
\newblock Mindsets: A view from two eras.
\newblock \emph{Perspectives on Psychological science}, 14(3):481--496.

\bibitem[{Ekman et~al.(1990)Ekman, Davidson, and Friesen}]{ekman1990duchenne}
Paul Ekman, Richard~J Davidson, and Wallace~V Friesen. 1990.
\newblock The duchenne smile: emotional expression and brain physiology: Ii.
\newblock \emph{Journal of personality and social psychology}, 58(2):342.

\bibitem[{ElSherief et~al.(2021)ElSherief, Ziems, Muchlinski, Anupindi,
  Seybolt, De~Choudhury, and Yang}]{elsherief-etal-2021-latent}
Mai ElSherief, Caleb Ziems, David Muchlinski, Vaishnavi Anupindi, Jordyn
  Seybolt, Munmun De~Choudhury, and Diyi Yang. 2021.
\newblock \href {https://aclanthology.org/2021.emnlp-main.29} {Latent hatred: A
  benchmark for understanding implicit hate speech}.
\newblock In \emph{Proceedings of the 2021 Conference on Empirical Methods in
  Natural Language Processing}, pages 345--363, Online and Punta Cana,
  Dominican Republic. Association for Computational Linguistics.

\bibitem[{Emmons and Shelton(2002)}]{emmons2002gratitude}
Robert~A Emmons and Charles~M Shelton. 2002.
\newblock Gratitude and the science of positive psychology.
\newblock \emph{Handbook of positive psychology}, 18:459--471.

\bibitem[{Enright and Fitzgibbons(2000)}]{enright2000helping}
Robert~D Enright and Richard~P Fitzgibbons. 2000.
\newblock \emph{Helping clients forgive: An empirical guide for resolving anger
  and restoring hope.}
\newblock American Psychological Association.

\bibitem[{Erisen et~al.(2014)Erisen, Lodge, and Taber}]{erisen2014affective}
Cengiz Erisen, Milton Lodge, and Charles~S Taber. 2014.
\newblock Affective contagion in effortful political thinking.
\newblock \emph{Political Psychology}, 35(2):187--206.

\bibitem[{Filipowicz et~al.(2011)Filipowicz, Barsade, and
  Melwani}]{filipowicz2011understanding}
Allan Filipowicz, Sigal Barsade, and Shimul Melwani. 2011.
\newblock Understanding emotional transitions: the interpersonal consequences
  of changing emotions in negotiations.
\newblock \emph{Journal of personality and social psychology}, 101(3):541.

\bibitem[{Fredrickson(2000)}]{fredrickson2000extracting}
Barbara~L Fredrickson. 2000.
\newblock Extracting meaning from past affective experiences: The importance of
  peaks, ends, and specific emotions.
\newblock \emph{Cognition \& Emotion}, 14(4):577--606.

\bibitem[{Fredrickson and Losada(2005)}]{fredrickson2005positive}
Barbara~L Fredrickson and Marcial~F Losada. 2005.
\newblock Positive affect and the complex dynamics of human flourishing.
\newblock \emph{American psychologist}, 60(7):678.

\bibitem[{Fu et~al.(2018)Fu, Tan, Peng, Zhao, and Yan}]{fu2018style}
Zhenxin Fu, Xiaoye Tan, Nanyun Peng, Dongyan Zhao, and Rui Yan. 2018.
\newblock \href
  {https://www.aaai.org/ocs/index.php/AAAI/AAAI18/paper/view/17015} {Style
  transfer in text: Exploration and evaluation}.
\newblock In \emph{Proceedings of the Thirty-Second {AAAI} Conference on
  Artificial Intelligence, (AAAI-18), the 30th innovative Applications of
  Artificial Intelligence (IAAI-18), and the 8th {AAAI} Symposium on
  Educational Advances in Artificial Intelligence (EAAI-18), New Orleans,
  Louisiana, USA, February 2-7, 2018}, pages 663--670. {AAAI} Press.

\bibitem[{G{\"a}chter et~al.(2009)G{\"a}chter, Orzen, Renner, and
  Starmer}]{gachter2009experimental}
Simon G{\"a}chter, Henrik Orzen, Elke Renner, and Chris Starmer. 2009.
\newblock Are experimental economists prone to framing effects? a natural field
  experiment.
\newblock \emph{Journal of Economic Behavior \& Organization}, 70(3):443--446.

\bibitem[{Good et~al.(2003)Good, Aronson, and Inzlicht}]{good2003improving}
Catherine Good, Joshua Aronson, and Michael Inzlicht. 2003.
\newblock Improving adolescents' standardized test performance: An intervention
  to reduce the effects of stereotype threat.
\newblock \emph{Journal of Applied Developmental Psychology}, 24(6):645--662.

\bibitem[{Grice(1975)}]{grice1975logic}
Herbert~P Grice. 1975.
\newblock Logic and conversation.
\newblock In \emph{Speech acts}, pages 41--58. Brill.

\bibitem[{Grier et~al.(2010)Grier, Thomas, Paxson, and Zhang}]{grier2010spam}
Chris Grier, Kurt Thomas, Vern Paxson, and Michael Zhang. 2010.
\newblock @ spam: the underground on 140 characters or less.
\newblock In \emph{Proceedings of the 17th ACM conference on Computer and
  communications security}, pages 27--37.

\bibitem[{Gross(2013)}]{gross2013handbook}
James~J Gross. 2013.
\newblock \emph{Handbook of emotion regulation}.
\newblock Guilford publications.

\bibitem[{Han et~al.(2018)Han, Zhu, Yu, Wang, Yao, Liu, and
  Sun}]{han2018fewrel}
Xu~Han, Hao Zhu, Pengfei Yu, Ziyun Wang, Yuan Yao, Zhiyuan Liu, and Maosong
  Sun. 2018.
\newblock \href {https://doi.org/10.18653/v1/D18-1514} {{F}ew{R}el: A
  large-scale supervised few-shot relation classification dataset with
  state-of-the-art evaluation}.
\newblock In \emph{Proceedings of the 2018 Conference on Empirical Methods in
  Natural Language Processing}, pages 4803--4809, Brussels, Belgium.
  Association for Computational Linguistics.

\bibitem[{Harris et~al.(2007)Harris, Thoresen, and
  Lopez}]{harris2007integrating}
Alex~HS Harris, Carl~E Thoresen, and Shane~J Lopez. 2007.
\newblock Integrating positive psychology into counseling: Why and (when
  appropriate) how.
\newblock \emph{Journal of Counseling \& Development}, 85(1):3--13.

\bibitem[{Hatfield et~al.(1993)Hatfield, Cacioppo, and
  Rapson}]{hatfield1993emotional}
Elaine Hatfield, John~T Cacioppo, and Richard~L Rapson. 1993.
\newblock Emotional contagion.
\newblock \emph{Current directions in psychological science}, 2(3):96--100.

\bibitem[{He and McAuley(2016)}]{he2016ups}
Ruining He and Julian~J. McAuley. 2016.
\newblock \href {https://doi.org/10.1145/2872427.2883037} {Ups and downs:
  Modeling the visual evolution of fashion trends with one-class collaborative
  filtering}.
\newblock In \emph{Proceedings of the 25th International Conference on World
  Wide Web, {WWW} 2016, Montreal, Canada, April 11 - 15, 2016}, pages 507--517.
  {ACM}.

\bibitem[{Hochreiter and Schmidhuber(1997)}]{hochreiter1997long}
Sepp Hochreiter and J{\"u}rgen Schmidhuber. 1997.
\newblock Long short-term memory.
\newblock \emph{Neural computation}, 9(8):1735--1780.

\bibitem[{Hovy(1987)}]{hovy1987generating}
Eduard Hovy. 1987.
\newblock Generating natural language under pragmatic constraints.
\newblock \emph{Journal of Pragmatics}, 11:689--719.

\bibitem[{Hu et~al.(2017)Hu, Yang, Liang, Salakhutdinov, and
  Xing}]{hu2017toward}
Zhiting Hu, Zichao Yang, Xiaodan Liang, Ruslan Salakhutdinov, and Eric~P. Xing.
  2017.
\newblock \href {http://proceedings.mlr.press/v70/hu17e.html} {Toward
  controlled generation of text}.
\newblock In \emph{Proceedings of the 34th International Conference on Machine
  Learning, {ICML} 2017, Sydney, NSW, Australia, 6-11 August 2017}, volume~70
  of \emph{Proceedings of Machine Learning Research}, pages 1587--1596. {PMLR}.

\bibitem[{Huang et~al.(2020)Huang, Zhu, Xiong, Zhang, Hu, and
  Xu}]{huang2020cycle}
Yufang Huang, Wentao Zhu, Deyi Xiong, Yiye Zhang, Changjian Hu, and Feiyu Xu.
  2020.
\newblock \href {https://doi.org/10.18653/v1/2020.coling-main.201}
  {Cycle-consistent adversarial autoencoders for unsupervised text style
  transfer}.
\newblock In \emph{Proceedings of the 28th International Conference on
  Computational Linguistics}, pages 2213--2223, Barcelona, Spain (Online).
  International Committee on Computational Linguistics.

\bibitem[{Jhamtani et~al.(2017)Jhamtani, Gangal, Hovy, and
  Nyberg}]{jhamtani2017shakespearizing}
Harsh Jhamtani, Varun Gangal, Eduard Hovy, and Eric Nyberg. 2017.
\newblock \href {https://doi.org/10.18653/v1/W17-4902} {Shakespearizing modern
  language using copy-enriched sequence to sequence models}.
\newblock In \emph{Proceedings of the Workshop on Stylistic Variation}, pages
  10--19, Copenhagen, Denmark. Association for Computational Linguistics.

\bibitem[{Jin et~al.(2020)Jin, Jin, Hu, Vechtomova, and Mihalcea}]{jin2020deep}
Di~Jin, Zhijing Jin, Zhiting Hu, Olga Vechtomova, and Rada Mihalcea. 2020.
\newblock \href {https://arxiv.org/abs/2011.00416} {Deep learning for text
  style transfer: A survey}.
\newblock \emph{ArXiv preprint}, abs/2011.00416.

\bibitem[{Jin et~al.(2019)Jin, Jin, Mueller, Matthews, and
  Santus}]{jin2019imat}
Zhijing Jin, Di~Jin, Jonas Mueller, Nicholas Matthews, and Enrico Santus. 2019.
\newblock \href {https://doi.org/10.18653/v1/D19-1306} {{IM}a{T}: Unsupervised
  text attribute transfer via iterative matching and translation}.
\newblock In \emph{Proceedings of the 2019 Conference on Empirical Methods in
  Natural Language Processing and the 9th International Joint Conference on
  Natural Language Processing (EMNLP-IJCNLP)}, pages 3097--3109, Hong Kong,
  China. Association for Computational Linguistics.

\bibitem[{Job et~al.(2015)Job, Walton, Bernecker, and
  Dweck}]{Job2015ImplicitTA}
V.~Job, G.~Walton, K.~Bernecker, and C.~Dweck. 2015.
\newblock Implicit theories about willpower predict self-regulation and grades
  in everyday life.
\newblock \emph{Journal of personality and social psychology}, 108 4:637--47.

\bibitem[{Job et~al.(2010)Job, Dweck, and Walton}]{job2010ego}
Veronika Job, Carol~S Dweck, and Gregory~M Walton. 2010.
\newblock Ego depletion -- is it all in your head? implicit theories about
  willpower affect self-regulation.
\newblock \emph{Psychological science}, 21(11):1686--1693.

\bibitem[{Johnson(2009)}]{johnson2009you}
Stefanie~K Johnson. 2009.
\newblock Do you feel what i feel? mood contagion and leadership outcomes.
\newblock \emph{The Leadership Quarterly}, 20(5):814--827.

\bibitem[{Kim et~al.(2020)Kim, Kim, and Kim}]{kim2020will}
Hyunwoo Kim, Byeongchang Kim, and Gunhee Kim. 2020.
\newblock \href {https://doi.org/10.18653/v1/2020.emnlp-main.65} {Will {I}
  sound like me? improving persona consistency in dialogues through pragmatic
  self-consciousness}.
\newblock In \emph{Proceedings of the 2020 Conference on Empirical Methods in
  Natural Language Processing (EMNLP)}, pages 904--916, Online. Association for
  Computational Linguistics.

\bibitem[{Korotkova et~al.(2019)Korotkova, Luhtaru, Del, Liin, Deksne, and
  Fishel}]{korotkova2019grammatical}
Elizaveta Korotkova, Agnes Luhtaru, Maksym Del, Krista Liin, Daiga Deksne, and
  Mark Fishel. 2019.
\newblock \href {https://arxiv.org/abs/1903.11283} {Grammatical error
  correction and style transfer via zero-shot monolingual translation}.
\newblock \emph{ArXiv preprint}, abs/1903.11283.

\bibitem[{Krause et~al.(2020)Krause, Gotmare, McCann, Keskar, Joty, Socher, and
  Rajani}]{krause2020gedi}
Ben Krause, Akhilesh~Deepak Gotmare, Bryan McCann, Nitish~Shirish Keskar,
  Shafiq Joty, Richard Socher, and Nazneen~Fatema Rajani. 2020.
\newblock \href {https://arxiv.org/abs/2009.06367} {Gedi: Generative
  discriminator guided sequence generation}.
\newblock \emph{ArXiv preprint}, abs/2009.06367.

\bibitem[{Lambert et~al.(2012)Lambert, Fincham, and
  Stillman}]{lambert2012gratitude}
Nathaniel~M Lambert, Frank~D Fincham, and Tyler~F Stillman. 2012.
\newblock Gratitude and depressive symptoms: The role of positive reframing and
  positive emotion.
\newblock \emph{Cognition \& emotion}, 26(4):615--633.

\bibitem[{Lample et~al.(2018)Lample, Conneau, Denoyer, and
  Ranzato}]{lample2018unsupervised}
Guillaume Lample, Alexis Conneau, Ludovic Denoyer, and Marc'Aurelio Ranzato.
  2018.
\newblock \href {https://openreview.net/forum?id=rkYTTf-AZ} {Unsupervised
  machine translation using monolingual corpora only}.
\newblock In \emph{6th International Conference on Learning Representations,
  {ICLR} 2018, Vancouver, BC, Canada, April 30 - May 3, 2018, Conference Track
  Proceedings}. OpenReview.net.

\bibitem[{Lample et~al.(2019)Lample, Subramanian, Smith, Denoyer, Ranzato, and
  Boureau}]{lample2018multiple}
Guillaume Lample, Sandeep Subramanian, Eric~Michael Smith, Ludovic Denoyer,
  Marc'Aurelio Ranzato, and Y{-}Lan Boureau. 2019.
\newblock \href {https://openreview.net/forum?id=H1g2NhC5KQ}
  {Multiple-attribute text rewriting}.
\newblock In \emph{7th International Conference on Learning Representations,
  {ICLR} 2019, New Orleans, LA, USA, May 6-9, 2019}. OpenReview.net.

\bibitem[{Lewis et~al.(2020)Lewis, Liu, Goyal, Ghazvininejad, Mohamed, Levy,
  Stoyanov, and Zettlemoyer}]{lewis2020bart}
Mike Lewis, Yinhan Liu, Naman Goyal, Marjan Ghazvininejad, Abdelrahman Mohamed,
  Omer Levy, Veselin Stoyanov, and Luke Zettlemoyer. 2020.
\newblock \href {https://doi.org/10.18653/v1/2020.acl-main.703} {{BART}:
  Denoising sequence-to-sequence pre-training for natural language generation,
  translation, and comprehension}.
\newblock In \emph{Proceedings of the 58th Annual Meeting of the Association
  for Computational Linguistics}, pages 7871--7880, Online. Association for
  Computational Linguistics.

\bibitem[{Li et~al.(2019)Li, Zhang, Gan, Cheng, Brockett, Dolan, and
  Sun}]{li-etal-2019-domain}
Dianqi Li, Yizhe Zhang, Zhe Gan, Yu~Cheng, Chris Brockett, Bill Dolan, and
  Ming-Ting Sun. 2019.
\newblock \href {https://doi.org/10.18653/v1/D19-1325} {Domain adaptive text
  style transfer}.
\newblock In \emph{Proceedings of the 2019 Conference on Empirical Methods in
  Natural Language Processing and the 9th International Joint Conference on
  Natural Language Processing (EMNLP-IJCNLP)}, pages 3304--3313, Hong Kong,
  China. Association for Computational Linguistics.

\bibitem[{Li et~al.(2018)Li, Jia, He, and Liang}]{li2018delete}
Juncen Li, Robin Jia, He~He, and Percy Liang. 2018.
\newblock \href {https://doi.org/10.18653/v1/N18-1169} {Delete, retrieve,
  generate: a simple approach to sentiment and style transfer}.
\newblock In \emph{Proceedings of the 2018 Conference of the North {A}merican
  Chapter of the Association for Computational Linguistics: Human Language
  Technologies, Volume 1 (Long Papers)}, pages 1865--1874, New Orleans,
  Louisiana. Association for Computational Linguistics.

\bibitem[{Lin(2004)}]{lin2004rouge}
Chin-Yew Lin. 2004.
\newblock \href {https://aclanthology.org/W04-1013} {{ROUGE}: A package for
  automatic evaluation of summaries}.
\newblock In \emph{Text Summarization Branches Out}, pages 74--81, Barcelona,
  Spain. Association for Computational Linguistics.

\bibitem[{Liu et~al.(2019)Liu, Ott, Goyal, Du, Joshi, Chen, Levy, Lewis,
  Zettlemoyer, and Stoyanov}]{liu2019roberta}
Yinhan Liu, Myle Ott, Naman Goyal, Jingfei Du, Mandar Joshi, Danqi Chen, Omer
  Levy, Mike Lewis, Luke Zettlemoyer, and Veselin Stoyanov. 2019.
\newblock \href {https://arxiv.org/abs/1907.11692} {Roberta: A robustly
  optimized bert pretraining approach}.
\newblock \emph{ArXiv preprint}, abs/1907.11692.

\bibitem[{Logel et~al.(2009)Logel, Iserman, Davies, Quinn, and
  Spencer}]{logel2009perils}
Christine Logel, Emma~C Iserman, Paul~G Davies, Diane~M Quinn, and Steven~J
  Spencer. 2009.
\newblock The perils of double consciousness: The role of thought suppression
  in stereotype threat.
\newblock \emph{Journal of Experimental Social Psychology}, 45(2):299--312.

\bibitem[{Logeswaran et~al.(2018)Logeswaran, Lee, and
  Bengio}]{logeswaran2018content}
Lajanugen Logeswaran, Honglak Lee, and Samy Bengio. 2018.
\newblock \href
  {https://proceedings.neurips.cc/paper/2018/hash/7cf64379eb6f29a4d25c4b6a2df713e4-Abstract.html}
  {Content preserving text generation with attribute controls}.
\newblock In \emph{Advances in Neural Information Processing Systems 31: Annual
  Conference on Neural Information Processing Systems 2018, NeurIPS 2018,
  December 3-8, 2018, Montr{\'{e}}al, Canada}, pages 5108--5118.

\bibitem[{Loria(2018)}]{loria2018textblob}
Steven Loria. 2018.
\newblock textblob documentation.
\newblock \emph{Release 0.15}, 2:269.

\bibitem[{Luo et~al.(2019)Luo, Li, Yang, Zhou, Tan, Chang, Sui, and
  Sun}]{luo2019towards}
Fuli Luo, Peng Li, Pengcheng Yang, Jie Zhou, Yutong Tan, Baobao Chang, Zhifang
  Sui, and Xu~Sun. 2019.
\newblock \href {https://doi.org/10.18653/v1/P19-1194} {Towards fine-grained
  text sentiment transfer}.
\newblock In \emph{Proceedings of the 57th Annual Meeting of the Association
  for Computational Linguistics}, pages 2013--2022, Florence, Italy.
  Association for Computational Linguistics.

\bibitem[{Ma et~al.(2020)Ma, Sap, Rashkin, and Choi}]{ma2020powertransformer}
Xinyao Ma, Maarten Sap, Hannah Rashkin, and Yejin Choi. 2020.
\newblock \href {https://doi.org/10.18653/v1/2020.emnlp-main.602}
  {{P}ower{T}ransformer: Unsupervised controllable revision for biased language
  correction}.
\newblock In \emph{Proceedings of the 2020 Conference on Empirical Methods in
  Natural Language Processing (EMNLP)}, pages 7426--7441, Online. Association
  for Computational Linguistics.

\bibitem[{Madaan et~al.(2020)Madaan, Setlur, Parekh, Poczos, Neubig, Yang,
  Salakhutdinov, Black, and Prabhumoye}]{madaan2020politeness}
Aman Madaan, Amrith Setlur, Tanmay Parekh, Barnabas Poczos, Graham Neubig,
  Yiming Yang, Ruslan Salakhutdinov, Alan~W Black, and Shrimai Prabhumoye.
  2020.
\newblock \href {https://doi.org/10.18653/v1/2020.acl-main.169} {Politeness
  transfer: A tag and generate approach}.
\newblock In \emph{Proceedings of the 58th Annual Meeting of the Association
  for Computational Linguistics}, pages 1869--1881, Online. Association for
  Computational Linguistics.

\bibitem[{Malmi et~al.(2020)Malmi, Severyn, and Rothe}]{malmi2020unsupervised}
Eric Malmi, Aliaksei Severyn, and Sascha Rothe. 2020.
\newblock \href {https://doi.org/10.18653/v1/2020.emnlp-main.699} {Unsupervised
  text style transfer with padded masked language models}.
\newblock In \emph{Proceedings of the 2020 Conference on Empirical Methods in
  Natural Language Processing (EMNLP)}, pages 8671--8680, Online. Association
  for Computational Linguistics.

\bibitem[{Martens et~al.(2006)Martens, Johns, Greenberg, and
  Schimel}]{martens2006combating}
Andy Martens, Michael Johns, Jeff Greenberg, and Jeff Schimel. 2006.
\newblock Combating stereotype threat: The effect of self-affirmation on
  women’s intellectual performance.
\newblock \emph{Journal of Experimental Social Psychology}, 42(2):236--243.

\bibitem[{Masters(1992)}]{masters1992use}
Mark~A Masters. 1992.
\newblock The use of positive reframing in the context of supervision.
\newblock \emph{Journal of Counseling \& Development}, 70(3):387--390.

\bibitem[{McDonald and Pustejovsky(1985)}]{mcdonald1985computational}
David~D. McDonald and James~D. Pustejovsky. 1985.
\newblock \href {https://aclanthology.org/E85-1027} {A computational theory of
  prose style for natural language generation}.
\newblock In \emph{Second Conference of the {E}uropean Chapter of the
  Association for Computational Linguistics}, Geneva, Switzerland. Association
  for Computational Linguistics.

\bibitem[{Mir et~al.(2019)Mir, Felbo, Obradovich, and
  Rahwan}]{mir2019evaluating}
Remi Mir, Bjarke Felbo, Nick Obradovich, and Iyad Rahwan. 2019.
\newblock \href {https://doi.org/10.18653/v1/N19-1049} {Evaluating style
  transfer for text}.
\newblock In \emph{Proceedings of the 2019 Conference of the North {A}merican
  Chapter of the Association for Computational Linguistics: Human Language
  Technologies, Volume 1 (Long and Short Papers)}, pages 495--504, Minneapolis,
  Minnesota. Association for Computational Linguistics.

\bibitem[{Muran and Motta(1993)}]{muran1993cognitive}
Elizabeth~M Muran and Robert~W Motta. 1993.
\newblock Cognitive distortions and irrational beliefs in post-traumatic
  stress, anxiety, and depressive disorders.
\newblock \emph{Journal of Clinical Psychology}, 49(2):166--176.

\bibitem[{Nalabandian and Ireland(2019)}]{nalabandian-ireland-2019-depressed}
Taleen Nalabandian and Molly Ireland. 2019.
\newblock \href {https://doi.org/10.18653/v1/W19-3008} {Depressed individuals
  use negative self-focused language when recalling recent interactions with
  close romantic partners but not family or {F}riends}.
\newblock In \emph{Proceedings of the Sixth Workshop on Computational
  Linguistics and Clinical Psychology}, pages 62--73, Minneapolis, Minnesota.
  Association for Computational Linguistics.

\bibitem[{Nie et~al.(2019)Nie, Yao, Wang, Pan, and Lin}]{nie2019simple}
Feng Nie, Jin-Ge Yao, Jinpeng Wang, Rong Pan, and Chin-Yew Lin. 2019.
\newblock \href {https://doi.org/10.18653/v1/P19-1256} {A simple recipe towards
  reducing hallucination in neural surface realisation}.
\newblock In \emph{Proceedings of the 57th Annual Meeting of the Association
  for Computational Linguistics}, pages 2673--2679, Florence, Italy.
  Association for Computational Linguistics.

\bibitem[{Niu et~al.(2018)Niu, Rao, and Carpuat}]{niu2018multi}
Xing Niu, Sudha Rao, and Marine Carpuat. 2018.
\newblock \href {https://aclanthology.org/C18-1086} {Multi-task neural models
  for translating between styles within and across languages}.
\newblock In \emph{Proceedings of the 27th International Conference on
  Computational Linguistics}, pages 1008--1021, Santa Fe, New Mexico, USA.
  Association for Computational Linguistics.

\bibitem[{Oshio(2012)}]{oshio2012all}
Atsushi Oshio. 2012.
\newblock An all-or-nothing thinking turns into darkness: Relations between
  dichotomous thinking and personality disorders 1.
\newblock \emph{Japanese Psychological Research}, 54(4):424--429.

\bibitem[{Papineni et~al.(2002)Papineni, Roukos, Ward, and
  Zhu}]{papineni2002bleu}
Kishore Papineni, Salim Roukos, Todd Ward, and Wei-Jing Zhu. 2002.
\newblock \href {https://doi.org/10.3115/1073083.1073135} {{B}leu: a method for
  automatic evaluation of machine translation}.
\newblock In \emph{Proceedings of the 40th Annual Meeting of the Association
  for Computational Linguistics}, pages 311--318, Philadelphia, Pennsylvania,
  USA. Association for Computational Linguistics.

\bibitem[{Prabhumoye et~al.(2018)Prabhumoye, Tsvetkov, Salakhutdinov, and
  Black}]{prabhumoye2018style}
Shrimai Prabhumoye, Yulia Tsvetkov, Ruslan Salakhutdinov, and Alan~W Black.
  2018.
\newblock \href {https://doi.org/10.18653/v1/P18-1080} {Style transfer through
  back-translation}.
\newblock In \emph{Proceedings of the 56th Annual Meeting of the Association
  for Computational Linguistics (Volume 1: Long Papers)}, pages 866--876,
  Melbourne, Australia. Association for Computational Linguistics.

\bibitem[{Pryzant et~al.(2020)Pryzant, Martinez, Dass, Kurohashi, Jurafsky, and
  Yang}]{pryzant2020automatically}
Reid Pryzant, Richard~Diehl Martinez, Nathan Dass, Sadao Kurohashi, Dan
  Jurafsky, and Diyi Yang. 2020.
\newblock Automatically neutralizing subjective bias in text.
\newblock In \emph{Proceedings of the aaai conference on artificial
  intelligence}, volume~34, pages 480--489.

\bibitem[{Radford et~al.(2019)Radford, Wu, Child, Luan, Amodei, and
  Sutskever}]{radford2019language}
Alec Radford, Jeffrey Wu, Rewon Child, David Luan, Dario Amodei, and Ilya
  Sutskever. 2019.
\newblock Language models are unsupervised multitask learners.
\newblock \emph{OpenAI blog}, 1(8):9.

\bibitem[{Raffel et~al.(2020)Raffel, Shazeer, Roberts, Lee, Narang, Matena,
  Zhou, Li, and Liu}]{raffel2020exploring}
Colin Raffel, Noam Shazeer, Adam Roberts, Katherine Lee, Sharan Narang, Michael
  Matena, Yanqi Zhou, Wei Li, and Peter~J Liu. 2020.
\newblock Exploring the limits of transfer learning with a unified text-to-text
  transformer.
\newblock \emph{Journal of Machine Learning Research}, 21(140):1--67.

\bibitem[{Rao and Tetreault(2018)}]{rao2018dear}
Sudha Rao and Joel Tetreault. 2018.
\newblock \href {https://doi.org/10.18653/v1/N18-1012} {Dear sir or madam, may
  {I} introduce the {GYAFC} dataset: Corpus, benchmarks and metrics for
  formality style transfer}.
\newblock In \emph{Proceedings of the 2018 Conference of the North {A}merican
  Chapter of the Association for Computational Linguistics: Human Language
  Technologies, Volume 1 (Long Papers)}, pages 129--140, New Orleans,
  Louisiana. Association for Computational Linguistics.

\bibitem[{Reimers and Gurevych(2019)}]{reimers-2019-sentence-bert}
Nils Reimers and Iryna Gurevych. 2019.
\newblock \href {https://doi.org/10.18653/v1/D19-1410} {Sentence-{BERT}:
  Sentence embeddings using {S}iamese {BERT}-networks}.
\newblock In \emph{Proceedings of the 2019 Conference on Empirical Methods in
  Natural Language Processing and the 9th International Joint Conference on
  Natural Language Processing (EMNLP-IJCNLP)}, pages 3982--3992, Hong Kong,
  China. Association for Computational Linguistics.

\bibitem[{Scheier et~al.(2001)Scheier, Carver, and
  Bridges}]{scheier2001optimism}
Michael~F Scheier, Charles~S Carver, and Michael~W Bridges. 2001.
\newblock Optimism, pessimism, and psychological well-being.

\bibitem[{Sears and Kraus(2009)}]{sears2009think}
Sharon Sears and Sue Kraus. 2009.
\newblock I think therefore i om: Cognitive distortions and coping style as
  mediators for the effects of mindfulness meditation on anxiety, positive and
  negative affect, and hope.
\newblock \emph{Journal of clinical psychology}, 65(6):561--573.

\bibitem[{See et~al.(2017)See, Liu, and Manning}]{see2017get}
Abigail See, Peter~J. Liu, and Christopher~D. Manning. 2017.
\newblock \href {https://doi.org/10.18653/v1/P17-1099} {Get to the point:
  Summarization with pointer-generator networks}.
\newblock In \emph{Proceedings of the 55th Annual Meeting of the Association
  for Computational Linguistics (Volume 1: Long Papers)}, pages 1073--1083,
  Vancouver, Canada. Association for Computational Linguistics.

\bibitem[{Shang et~al.(2019)Shang, Li, Fu, Bing, Zhao, Shi, and
  Yan}]{shang2019semi}
Mingyue Shang, Piji Li, Zhenxin Fu, Lidong Bing, Dongyan Zhao, Shuming Shi, and
  Rui Yan. 2019.
\newblock \href {https://doi.org/10.18653/v1/D19-1499} {Semi-supervised text
  style transfer: Cross projection in latent space}.
\newblock In \emph{Proceedings of the 2019 Conference on Empirical Methods in
  Natural Language Processing and the 9th International Joint Conference on
  Natural Language Processing (EMNLP-IJCNLP)}, pages 4937--4946, Hong Kong,
  China. Association for Computational Linguistics.

\bibitem[{Sheehan(2018)}]{sheehan2018crowdsourcing}
Kim~Bartel Sheehan. 2018.
\newblock Crowdsourcing research: data collection with amazon’s mechanical
  turk.
\newblock \emph{Communication Monographs}, 85(1):140--156.

\bibitem[{Shen et~al.(2019)Shen, Fried, Andreas, and
  Klein}]{shen2019pragmatically}
Sheng Shen, Daniel Fried, Jacob Andreas, and Dan Klein. 2019.
\newblock \href {https://doi.org/10.18653/v1/N19-1410} {Pragmatically
  informative text generation}.
\newblock In \emph{Proceedings of the 2019 Conference of the North {A}merican
  Chapter of the Association for Computational Linguistics: Human Language
  Technologies, Volume 1 (Long and Short Papers)}, pages 4060--4067,
  Minneapolis, Minnesota. Association for Computational Linguistics.

\bibitem[{Shen et~al.(2017)Shen, Lei, Barzilay, and Jaakkola}]{shen2017style}
Tianxiao Shen, Tao Lei, Regina Barzilay, and Tommi~S. Jaakkola. 2017.
\newblock \href
  {https://proceedings.neurips.cc/paper/2017/hash/2d2c8394e31101a261abf1784302bf75-Abstract.html}
  {Style transfer from non-parallel text by cross-alignment}.
\newblock In \emph{Advances in Neural Information Processing Systems 30: Annual
  Conference on Neural Information Processing Systems 2017, December 4-9, 2017,
  Long Beach, CA, {USA}}, pages 6830--6841.

\bibitem[{Sherman et~al.(2009)Sherman, Cohen, Nelson, Nussbaum, Bunyan, and
  Garcia}]{sherman2009affirmed}
David~K Sherman, Geoffrey~L Cohen, Leif~D Nelson, A~David Nussbaum, Debra~P
  Bunyan, and Julio Garcia. 2009.
\newblock Affirmed yet unaware: exploring the role of awareness in the process
  of self-affirmation.
\newblock \emph{Journal of personality and social psychology}, 97(5):745.

\bibitem[{Shickel et~al.(2020)Shickel, Siegel, Heesacker, Benton, and
  Rashidi}]{shickel2020automatic}
Benjamin Shickel, Scott Siegel, Martin Heesacker, Sherry Benton, and Parisa
  Rashidi. 2020.
\newblock Automatic detection and classification of cognitive distortions in
  mental health text.
\newblock In \emph{2020 IEEE 20th International Conference on Bioinformatics
  and Bioengineering (BIBE)}, pages 275--280. IEEE.

\bibitem[{Silverman et~al.(2013)Silverman, Logel, and
  Cohen}]{silverman2013self}
Arielle Silverman, Christine Logel, and Geoffrey~L Cohen. 2013.
\newblock Self-affirmation as a deliberate coping strategy: The moderating role
  of choice.
\newblock \emph{Journal of Experimental Social Psychology}, 49(1):93--98.

\bibitem[{Simms et~al.(2017)Simms, Ramstedt, Rich, Richards, Martinez, and
  Giraud-Carrier}]{simms2017detecting}
Taetem Simms, Clayton Ramstedt, Megan Rich, Michael Richards, T~Martinez, and
  C~Giraud-Carrier. 2017.
\newblock Detecting cognitive distortions through machine learning text
  analytics.
\newblock In \emph{2017 IEEE international conference on healthcare informatics
  (ICHI)}, pages 508--512. IEEE.

\bibitem[{Steele(1988)}]{steele1988psychology}
Claude~M Steele. 1988.
\newblock The psychology of self-affirmation: Sustaining the integrity of the
  self.
\newblock In \emph{Advances in experimental social psychology}, volume~21,
  pages 261--302. Elsevier.

\bibitem[{Sudhakar et~al.(2019)Sudhakar, Upadhyay, and
  Maheswaran}]{sudhakar2019transforming}
Akhilesh Sudhakar, Bhargav Upadhyay, and Arjun Maheswaran. 2019.
\newblock \href {https://doi.org/10.18653/v1/D19-1322} {{``}transforming{''}
  delete, retrieve, generate approach for controlled text style transfer}.
\newblock In \emph{Proceedings of the 2019 Conference on Empirical Methods in
  Natural Language Processing and the 9th International Joint Conference on
  Natural Language Processing (EMNLP-IJCNLP)}, pages 3269--3279, Hong Kong,
  China. Association for Computational Linguistics.

\bibitem[{Sullivan et~al.(2001)Sullivan, Rodgers, and
  Kirsch}]{sullivan2001catastrophizing}
Michael~JL Sullivan, Wendy~M Rodgers, and Irving Kirsch. 2001.
\newblock Catastrophizing, depression and expectancies for pain and emotional
  distress.
\newblock \emph{Pain}, 91(1-2):147--154.

\bibitem[{Sy and Choi(2013)}]{sy2013contagious}
Thomas Sy and Jin~Nam Choi. 2013.
\newblock Contagious leaders and followers: Exploring multi-stage mood
  contagion in a leader activation and member propagation (lamp) model.
\newblock \emph{Organizational Behavior and Human Decision Processes},
  122(2):127--140.

\bibitem[{Sy et~al.(2005)Sy, C{\^o}t{\'e}, and Saavedra}]{sy2005contagious}
Thomas Sy, St{\'e}phane C{\^o}t{\'e}, and Richard Saavedra. 2005.
\newblock The contagious leader: impact of the leader's mood on the mood of
  group members, group affective tone, and group processes.
\newblock \emph{Journal of applied psychology}, 90(2):295.

\bibitem[{Walton and Brady(2020)}]{walton2020bad}
Gregory~M Walton and Shannon~T Brady. 2020.
\newblock “bad” things reconsidered.
\newblock In \emph{Applications of social psychology}, pages 58--81. Routledge.

\bibitem[{Wang et~al.(2016)Wang, Chen, Rochford, and Qiang}]{wang2016text}
Tong Wang, Ping Chen, John Rochford, and Jipeng Qiang. 2016.
\newblock \href
  {http://www.aaai.org/ocs/index.php/AAAI/AAAI16/paper/view/11944} {Text
  simplification using neural machine translation}.
\newblock In \emph{Proceedings of the Thirtieth {AAAI} Conference on Artificial
  Intelligence, February 12-17, 2016, Phoenix, Arizona, {USA}}, pages
  4270--4271. {AAAI} Press.

\bibitem[{Watkins et~al.(2008)Watkins, Cruz, Holben, and
  Kolts}]{watkins2008taking}
Philip~C Watkins, Lilia Cruz, Heather Holben, and Russell~L Kolts. 2008.
\newblock Taking care of business? grateful processing of unpleasant memories.
\newblock \emph{The Journal of Positive Psychology}, 3(2):87--99.

\bibitem[{Wubben et~al.(2012)Wubben, van~den Bosch, and
  Krahmer}]{wubben2012sentence}
Sander Wubben, Antal van~den Bosch, and Emiel Krahmer. 2012.
\newblock \href {https://aclanthology.org/P12-1107} {Sentence simplification by
  monolingual machine translation}.
\newblock In \emph{Proceedings of the 50th Annual Meeting of the Association
  for Computational Linguistics (Volume 1: Long Papers)}, pages 1015--1024,
  Jeju Island, Korea. Association for Computational Linguistics.

\bibitem[{Xu et~al.(2016)Xu, Napoles, Pavlick, Chen, and
  Callison-Burch}]{xu2016optimizing}
Wei Xu, Courtney Napoles, Ellie Pavlick, Quanze Chen, and Chris Callison-Burch.
  2016.
\newblock \href {https://doi.org/10.1162/tacl_a_00107} {Optimizing statistical
  machine translation for text simplification}.
\newblock \emph{Transactions of the Association for Computational Linguistics},
  4:401--415.

\bibitem[{Xu et~al.(2012)Xu, Ritter, Dolan, Grishman, and
  Cherry}]{xu2012paraphrasing}
Wei Xu, Alan Ritter, Bill Dolan, Ralph Grishman, and Colin Cherry. 2012.
\newblock \href {https://aclanthology.org/C12-1177} {Paraphrasing for style}.
\newblock In \emph{Proceedings of {COLING} 2012}, pages 2899--2914, Mumbai,
  India. The COLING 2012 Organizing Committee.

\bibitem[{Yang et~al.(2019)Yang, Dai, Yang, Carbonell, Salakhutdinov, and
  Le}]{yang2019xlnet}
Zhilin Yang, Zihang Dai, Yiming Yang, Jaime~G. Carbonell, Ruslan Salakhutdinov,
  and Quoc~V. Le. 2019.
\newblock \href
  {https://proceedings.neurips.cc/paper/2019/hash/dc6a7e655d7e5840e66733e9ee67cc69-Abstract.html}
  {Xlnet: Generalized autoregressive pretraining for language understanding}.
\newblock In \emph{Advances in Neural Information Processing Systems 32: Annual
  Conference on Neural Information Processing Systems 2019, NeurIPS 2019,
  December 8-14, 2019, Vancouver, BC, Canada}, pages 5754--5764.

\bibitem[{Yeager et~al.(2014)Yeager, Johnson, Spitzer, Trzesniewski, Powers,
  and Dweck}]{yeager2014far}
David~Scott Yeager, Rebecca Johnson, Brian~James Spitzer, Kali~H Trzesniewski,
  Joseph Powers, and Carol~S Dweck. 2014.
\newblock The far-reaching effects of believing people can change: implicit
  theories of personality shape stress, health, and achievement during
  adolescence.
\newblock \emph{Journal of personality and social psychology}, 106(6):867.

\bibitem[{Zhang et~al.(2020{\natexlab{a}})Zhang, Kishore, Wu, Weinberger, and
  Artzi}]{zhang2019bertscore}
Tianyi Zhang, Varsha Kishore, Felix Wu, Kilian~Q. Weinberger, and Yoav Artzi.
  2020{\natexlab{a}}.
\newblock \href {https://openreview.net/forum?id=SkeHuCVFDr} {Bertscore:
  Evaluating text generation with {BERT}}.
\newblock In \emph{8th International Conference on Learning Representations,
  {ICLR} 2020, Addis Ababa, Ethiopia, April 26-30, 2020}. OpenReview.net.

\bibitem[{Zhang et~al.(2012)Zhang, Zhu, and Liang}]{zhang2012detecting}
Xianchao Zhang, Shaoping Zhu, and Wenxin Liang. 2012.
\newblock Detecting spam and promoting campaigns in the twitter social network.
\newblock In \emph{2012 IEEE 12th international conference on data mining},
  pages 1194--1199. IEEE.

\bibitem[{Zhang and Lapata(2017)}]{zhang2017sentence}
Xingxing Zhang and Mirella Lapata. 2017.
\newblock \href {https://doi.org/10.18653/v1/D17-1062} {Sentence simplification
  with deep reinforcement learning}.
\newblock In \emph{Proceedings of the 2017 Conference on Empirical Methods in
  Natural Language Processing}, pages 584--594, Copenhagen, Denmark.
  Association for Computational Linguistics.

\bibitem[{Zhang et~al.(2020{\natexlab{b}})Zhang, Ge, and
  Sun}]{zhang-etal-2020-parallel}
Yi~Zhang, Tao Ge, and Xu~Sun. 2020{\natexlab{b}}.
\newblock \href {https://doi.org/10.18653/v1/2020.acl-main.294} {Parallel data
  augmentation for formality style transfer}.
\newblock In \emph{Proceedings of the 58th Annual Meeting of the Association
  for Computational Linguistics}, pages 3221--3228, Online. Association for
  Computational Linguistics.

\bibitem[{Zhao et~al.(2018)Zhao, Kim, Zhang, Rush, and
  LeCun}]{zhao2018adversarially}
Junbo~Jake Zhao, Yoon Kim, Kelly Zhang, Alexander~M. Rush, and Yann LeCun.
  2018.
\newblock \href {http://proceedings.mlr.press/v80/zhao18b.html} {Adversarially
  regularized autoencoders}.
\newblock In \emph{Proceedings of the 35th International Conference on Machine
  Learning, {ICML} 2018, Stockholmsm{\"{a}}ssan, Stockholm, Sweden, July 10-15,
  2018}, volume~80 of \emph{Proceedings of Machine Learning Research}, pages
  5897--5906. {PMLR}.

\bibitem[{Zhu et~al.(2010)Zhu, Bernhard, and Gurevych}]{zhu2010monolingual}
Zhemin Zhu, Delphine Bernhard, and Iryna Gurevych. 2010.
\newblock \href {https://aclanthology.org/C10-1152} {A monolingual tree-based
  translation model for sentence simplification}.
\newblock In \emph{Proceedings of the 23rd International Conference on
  Computational Linguistics (Coling 2010)}, pages 1353--1361, Beijing, China.
  Coling 2010 Organizing Committee.

\end{thebibliography}

\appendix

\section{Data Quality-Control Methods}
\label{appdx:data_quality_control}
We used programmatic methods to ensure high-quality reframing annotations at submission time. Workers could not submit their task if the reframe: (1) contained fewer than 3 word types; (2) had a length less than 25\% of the original text; (3) had more than 3 repetitions of a single bigram; or (4) was too similar to the original text, with a token Jaccard Similarity greater than 90\%. Furthermore, we used the LanguageTool API\footnote{\url{api.languagetoolplus.com/v2/check}} to prompt workers to fix any grammatical mistakes in their writing. Cumulatively, these heuristics greatly improved the annotation quality. 
Later, in the post-processing stage, we employed additional programmatic measures as well as manual quality-checks to filter out the unsatisfactory examples. This process was iterated after each batch, with a batch size of 100. First, one of the authors manually checked any sentences where annotators had scored the \textit{original text} with a postivity score greater than 3 (out of 5). If that author found that the text was not negative enough or did not contain the requisite \textit{cognitive distortions} to warrant a substantial reframing, the sentence was removed from the corpus. Next, we considered all \textit{neutralizing} reframes with a score less than 4 (out of 5). If the text was not effectively neutralized, we removed the sentence from the corpus. Then we considered all annotations containing the first person pronoun \textit{you}. If the text abandoned the author's first-person voice and shifted into a 3rd-person critique or commentary (e.g. ``\textit{I feel hopeless}'' $\rightarrow$ ``\textit{you should find hope}''), then we removed this from the corpus. Finally, we grouped the annotations by Worker ID and, for each worker, scanned the top 10 
annotations. If the annotator produced poor quality work, we removed the examples and blocked the worker from future tasks. After a last pass through the data to manually correct noticeable punctuation and grammar errors, we were left with our cleaned corpus of 8,349 reframed sentences. 

\section{Task Interface}

\begin{figure}
\centering
\includegraphics[width =\columnwidth]{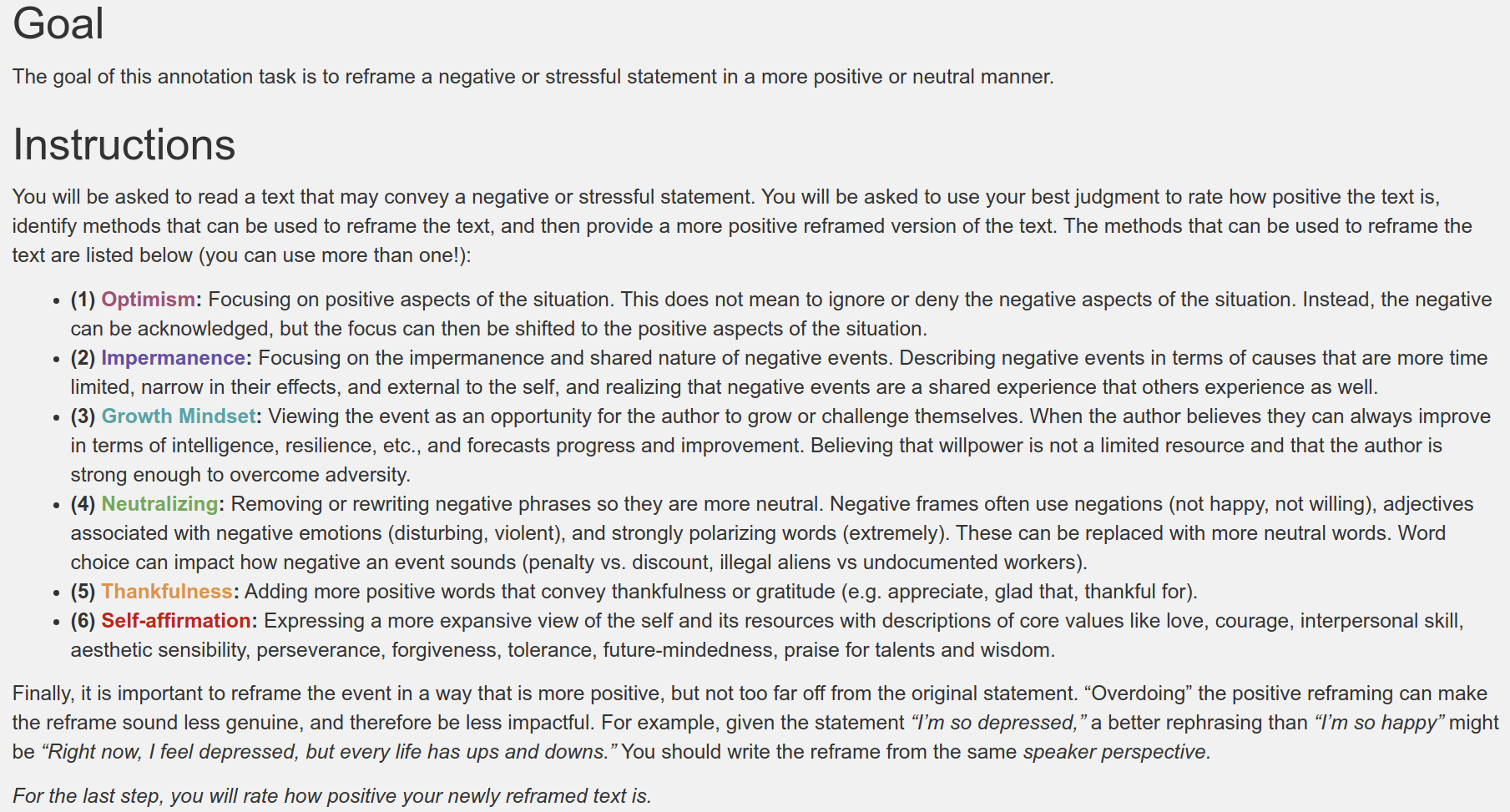}
\caption{Instructions for the Positive Reframing HIT.} 
\label{fig:task_instructions}
\end{figure}

\begin{figure}
\centering
\includegraphics[width =\columnwidth]{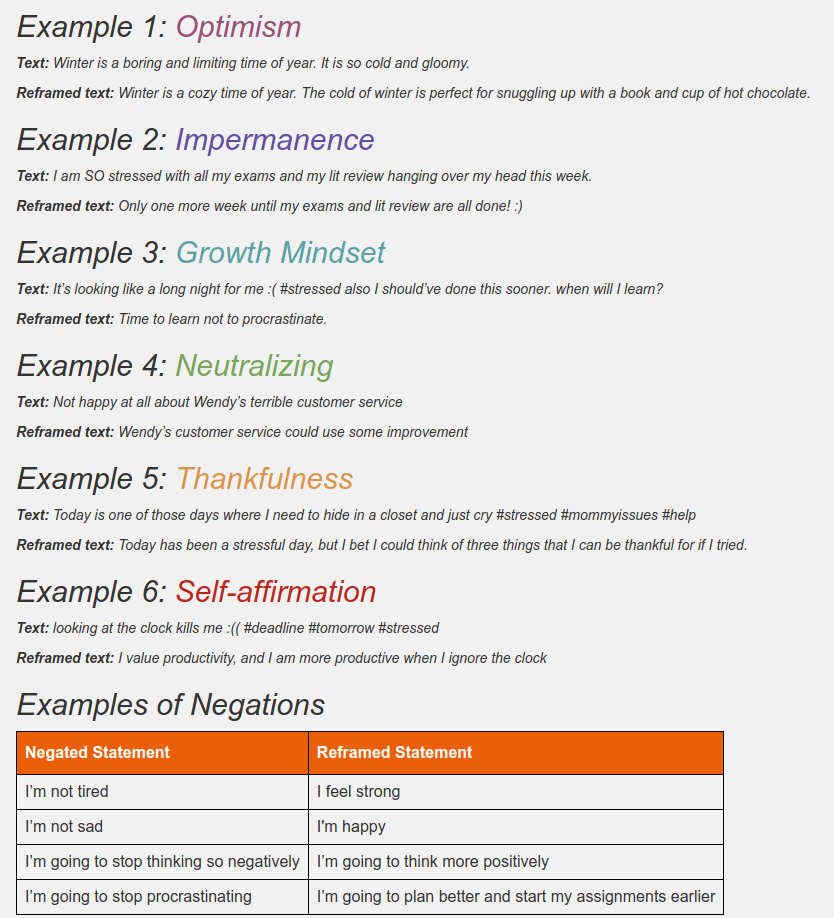}
\caption{Example reframes.} 
\label{fig:task_examples}
\end{figure}

\begin{figure*}
\centering
\includegraphics[width =\textwidth]{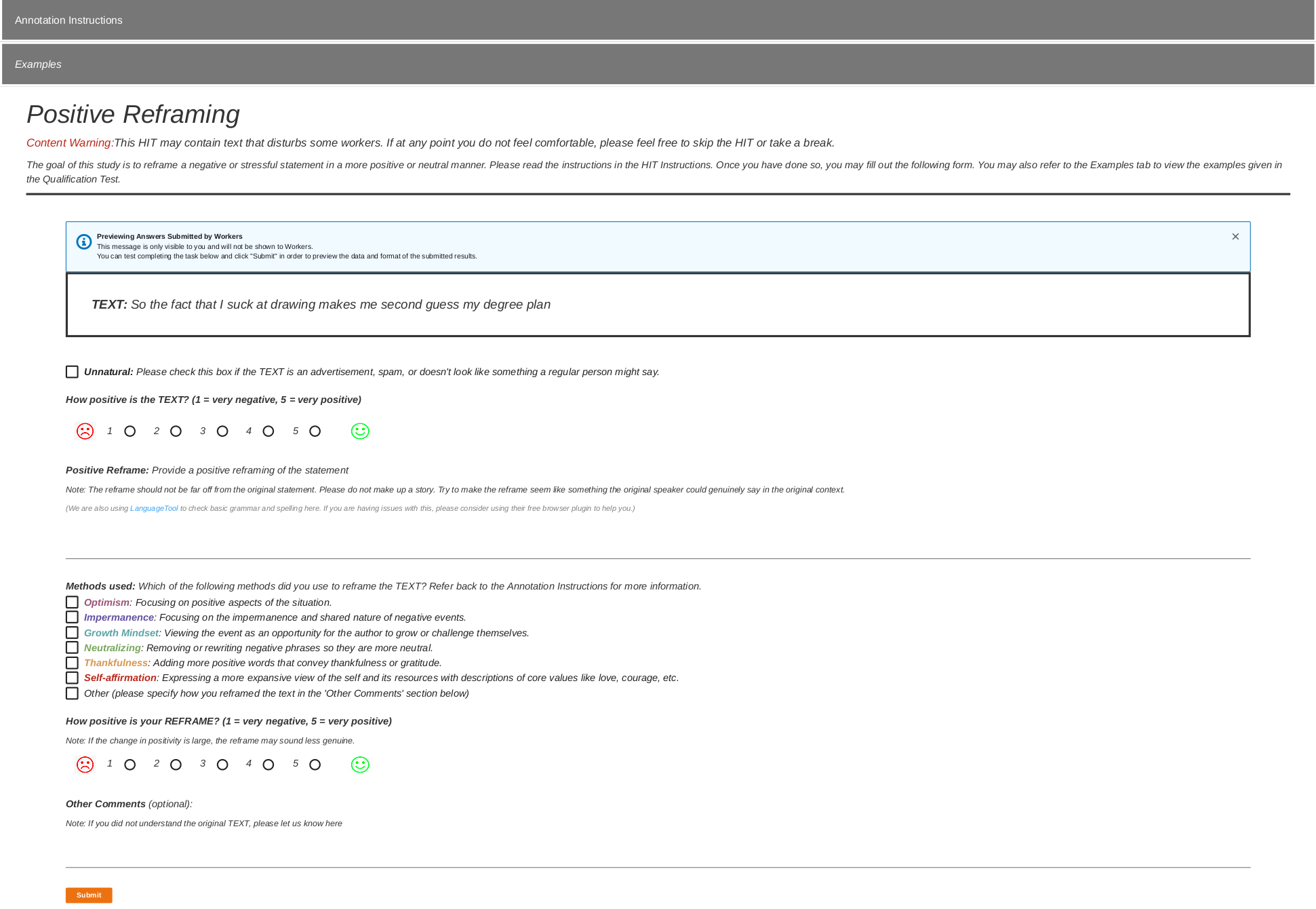}
\caption{Amazon Mechanical Turk interface used to collect positive reframes (in Section~\ref{subsec:annotation}).} 
\label{fig:task_interface}
\end{figure*}

\begin{figure*}
\centering
\includegraphics[width =\textwidth]{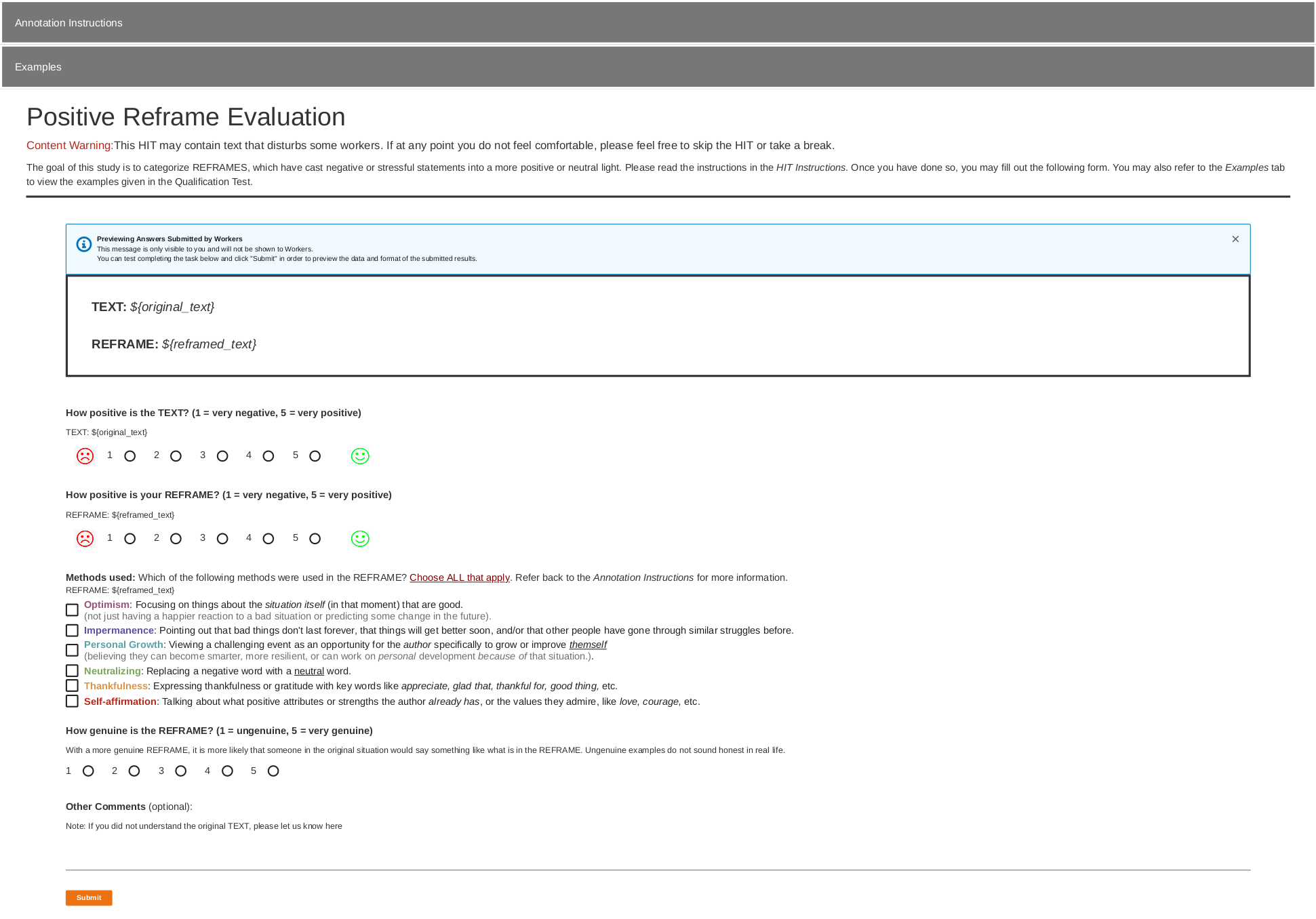}
\caption{Amazon Mechanical Turk interface used to find inter-annotator agreement for the taxonomy (in Section~\ref{subsec:data_quality}).} 
\label{fig:secondary_task_interface}
\end{figure*}

Figure~\ref{fig:task_instructions} shows the Instructions we gave to the Amazon Mechanical Turk (MTurk) workers. Figure~\ref{fig:task_examples} shows the examples we displayed for each reframe strategy. Figure~\ref{fig:task_interface} shows the MTurk HIT interface that we used for the Section~\ref{subsec:annotation} task to collect positive reframes with their associated strategies as well as the positivity scores for both the original TEXT and the REFRAME. Figure~\ref{fig:secondary_task_interface} shows the interface for the Section~\ref{subsec:data_quality} task where we collected new strategy labels for prior annotations to compute inter-annotator agreement scores.

\section{Few-shot Learning Setting}
\label{appdx:few_shot}
Following \cite{han2018fewrel,soares2019matching} and others, we consider 5-shot learning. We pull 5 representative exemplars from the training set to indicate a range of strategies: 

\dotfill
\begin{quote}
    \textsc{Negative}: "I have a huge project due tomorrow morning. But where do I have to be, a stupid basketball game dumb"

    \textsc{Positive}: "I should plan ahead next time so that my basketball game does not conflict too closely with my projects."
\end{quote}

\dotfill

\begin{quote}
    \textsc{Negative}: "This has been like the worst week ever im so done with everything. sick tired"
   
 \textsc{Positive}: "I made it to the end of the most challenging week ever!"
\end{quote}

\dotfill

\begin{quote}
    \textsc{Negative}: "Ugh my mac is starting to slow up and I need to figure out how to defragment the hard drive..."
    
 \textsc{Positive}: "I need to defragment the hard drive to speed up my mac. Good thing I'm smart, and I know I can do this."

\end{quote}

\dotfill

\begin{quote}
     \textsc{Negative}: "I am SO stressed with all my exams and my lit review hanging over my head this week."
     
 \textsc{Positive}: "Only one more week until my exams and lit review are all done!"

\end{quote}

\dotfill

\begin{quote}
     \textsc{Negative}: "I am the only person I know who writes a healthy grocery list and plans meals when I am stressed:( CantSleep"
     
 \textsc{Positive}: "I'm so thankful that I am still able to eat healthy even when I'm stressed."
\end{quote}

\dotfill

\section{Example Reframes}
\label{appdx:example_reframes}

In Table~\ref{tab:generation_model_comparison}, we compare examples of model-generated reframes from different models. The examples are structurally and semantically diverse, which may suggest that different architectures could serve as complementary systems in a broader effort to introduce a range of positive perspectives in text. However, the generations are not perfect. In this particular example, CopyNMT and GPT-2 fail to integrate the key concept of the ``heavy workload'' into their reframe generations. There is still significant room to improve upon these models in future work.

We were also interested in the success of the \textit{Controlled} generations $p(t, \boldsymbol{\psi}_t|s)$. Do models appropriately follow a given strategy when the generation is conditioned on that strategy? Table~\ref{tab:partial_inference} shows different reconstructions of the same text using different controlling strategies. Qualitatively, we can see that controlled generation is a reasonable method for influencing the most prominent strategy. The `thankfulness', `growth' and `self-affirmation' strategies are particularly distinct.

\begin{table*}[]
\resizebox{\textwidth}{!}{%
\begin{tabular}{|c|c|c|}
\hline
\multicolumn{2}{|c|}{\textbf{Original Text}}                   & \begin{tabular}[c]{@{}c@{}}Always stressing and thinking about loads of things at once\\  need I take it one at a time overload stressed need to rant.\end{tabular}                                                                 \\ \hline
\multicolumn{2}{|c|}{\textbf{Ground Truth}}                    & \begin{tabular}[c]{@{}c@{}}Loads of things on my mind, I need to make a list, \\ prioritise and work through it all calmly and I will feel much better.\end{tabular}                                                                \\ \hline
\multirow{9}{*}{\textbf{Model Generation}} & \textbf{Random}            & \begin{tabular}[c]{@{}c@{}}Annoyed because the USPS lost 1 of the 8 flat rate boxes of books \\ I shipped from California to Kentucky. Hope it shows up :-/\end{tabular}                                                            \\ \cline{2-3} 
                                           & \textbf{SBERT}             & \begin{tabular}[c]{@{}c@{}}Really stressed at the moment but I know that this will pass \\ and I'll get everything sorted out.\end{tabular}                                                                                         \\ \cline{2-3} 
                                           & \textbf{Seq2seq-LSTM}      & \begin{tabular}[c]{@{}c@{}}i have a lot of things to do it is a good night's sleep \\ and i can get through it.\end{tabular}                                                                                                        \\ \cline{2-3} 
                                           & \textbf{CopyNMT}           & \begin{tabular}[c]{@{}c@{}}Always stressing and thinking easy of things of things \\ at things at all, but I know that I need it.\end{tabular}                                                                                      \\ \cline{2-3} 
                                           & \textbf{GPT}               & \begin{tabular}[c]{@{}c@{}}i ' m so relieved that i ' m going to get to organize my thoughts \\ in such a short time ! i hope i am done \\ with the list of things i have to do today , so that i can go back to bed .\end{tabular} \\ \cline{2-3} 
                                           & \textbf{GPT-2 No-pretrain} & I'm going to try to seek a new job to get some newspect me through today.                                                                                                                                                           \\ \cline{2-3} 
                                           & \textbf{GPT-2}             & I should be more energetic in my thoughts, and not worry too much.                                                                                                                                                                  \\ \cline{2-3} 
                                           & \textbf{T5}                & \begin{tabular}[c]{@{}c@{}}I am stressed and thinking about loads of things at once, \\ I need to take it one at a time stressed need to rant. \\ I'm sure I'll be able to focus on it.\end{tabular}                                \\ \cline{2-3} 
                                           & \textbf{BART}              & \begin{tabular}[c]{@{}c@{}}I am stressed and thinking about loads of things at once, \\ but I am going to take it one at a time.\end{tabular}                                                                                       \\ \hline
\end{tabular}
}
\caption{\textbf{A model comparison} for reframing the same text. Here, we observe significant diversity in the generations, suggesting that these different architectures could serve as complementary systems in an effort to restruture and transform negative text on the web.}
\label{tab:generation_model_comparison}
\end{table*}

\begin{table*}[]
\resizebox{\textwidth}{!}{%
\begin{tabular}{@{}l|l|l@{}}
\toprule
\multicolumn{2}{l|}{\textbf{Original Text}}                             & I'm doing really well in some class and bad in other classes I hate this.                                                     \\ \midrule
\multirow{6}{*}{\textbf{GPT-2 Generations}} & \textbf{Growth Mindset}   & I'm doing really well in some class and bad in other classes I hate this. But I'm doing better than I was at the age of 11.   \\ \cmidrule(l){2-3} 
                                            & \textbf{Impermanence}     & I'm having a hard time in some classes and I hope to get through the rest of the semester.                                    \\ \cmidrule(l){2-3} 
                                            & \textbf{Neutralizing}     & I'm doing really well in some class and bad in other classes I hate this. But I'm going to go clear it out.                   \\ \cmidrule(l){2-3} 
                                            & \textbf{Optimism}         & I need to take a good hard look at myself and realize that I am quite capable of doing better in other classes.               \\ \cmidrule(l){2-3} 
                                            & \textbf{Self-affirmation} & I'm really happy about my class, I'm doing really well in some class, bad or not, I can do better in another.                 \\ \cmidrule(l){2-3} 
                                            & \textbf{Thankfulness}     & I'm so lucky that I'm able to get my classes to a good start and get off the couch.                                           \\ \midrule
\multirow{6}{*}{\textbf{T5 Generations}}    & \textbf{Growth Mindset}   & I'm doing really well in some class and bad in other classes I hate this, but I'll be able to do it again.                    \\ \cmidrule(l){2-3} 
                                            & \textbf{Impermanence}     & I'm doing really well in some class and bad in other classes I hate this. But I'll be able to do my best in the future.       \\ \cmidrule(l){2-3} 
                                            & \textbf{Neutralizing}     & I'm doing really well in some class and bad in other classes I hate this. But I can't wait to see if I can do it.             \\ \cmidrule(l){2-3} 
                                            & \textbf{Optimism}         & I'm doing really well in some class and bad in other classes I hate this. I hope I'll be able to do better in the future.     \\ \cmidrule(l){2-3} 
                                            & \textbf{Self-affirmation} & I'm doing really well in some class and bad in other classes I hate this. It's a good thing to do, but I'll be able to do it. \\ \cmidrule(l){2-3} 
                                            & \textbf{Thankfulness}     & I'm doing really well in some class and bad in other classes I hate this. But I can't wait to see if I can do it.             \\ \midrule
\multirow{6}{*}{\textbf{BART Generations}}  & \textbf{Growth Mindset}   & I'm doing really well in some class and bad in other classes. I'm going to try to improve my grades.                          \\ \cmidrule(l){2-3} 
                                            & \textbf{Impermanence}     & I'm doing really well in some class and bad in other classes, but I'm sure it will all work out in the end.                   \\ \cmidrule(l){2-3} 
                                            & \textbf{Neutralizing}     & I'm doing really well in some class and bad in other classes I don't like this.                                               \\ \cmidrule(l){2-3} 
                                            & \textbf{Optimism}         & I'm doing really well in some class and bad in other classes, but I'm sure it will all work out.                              \\ \cmidrule(l){2-3} 
                                            & \textbf{Self-affirmation} & I'm doing really well in some class and bad in other classes, but I know I can do better.                                     \\ \cmidrule(l){2-3} 
                                            & \textbf{Thankfulness}     & I'm doing really well in some class and bad in other classes, but I'm thankful that I have the opportunity to study.          \\ \bottomrule
\end{tabular}
}
\caption{\textbf{A model comparison} for reframing the same text using different controlling strategy. Here, we observe models can learn some information from the input strategy label and make distinctive generations, especially for the `thankfulness', `growth' and `self-affirmation' strategies.}
\label{tab:partial_inference}
\end{table*}

\end{document}